\documentclass[copyright,creativecommons]{eptcs}
\providecommand{\event}{GRAPHite 2012} 
\usepackage{graphicx,color}   
\usepackage{url}   
\usepackage{listings}
\usepackage{rotating}

\newcommand{\makemath}[1]{\relax\ifmmode #1\relax\else $#1$\ \fi}

\newcommand{\action}{\makemath{a}}
\newcommand{\Actions}{\makemath{{\cal A}}}

\newcommand{\legnone}{\includegraphics{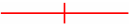}}
\newcommand{\legdisjvar}{\includegraphics{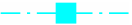}}
\newcommand{\legdisjite}{\includegraphics{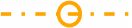}}
\newcommand{\legdisjgen}{\includegraphics{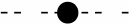}}
\newcommand{\legnodesspath}{\includegraphics{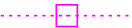}}
\newcommand{\legnodescomp}{\includegraphics{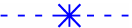}}
\newcommand{\legstateslex}{\includegraphics{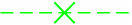}}

\usepackage{amsthm}
\newtheorem{theorem}{Theorem}
\newtheorem{definition}{Definition}

\usepackage{amssymb}
\usepackage[english]{babel}
\usepackage[autolanguage]{numprint}
\usepackage{subfig}
\usepackage{graphicx}
\graphicspath{{graphics/}}
\usepackage{color}
\usepackage{url}
\usepackage{tikz}
\usepackage{tikz-qtree}

\def\titlerunning{Lex-Partitioning}
\def\authorrunning{S. Edelkamp, P. Kissmann \& A. Torralba}
\title{Lex-Partitioning: A New Option for BDD Search}
\author{Stefan Edelkamp
\institute{Faculty of Mathematics and Computer Science \\
Universit{\"a}t Bremen, Germany\\
edelkamp@tzi.de}
\and
Peter Kissmann
\institute{Department of Computer Science \\
Universit{\"a}t des Saarlandes, Saarbr{\"u}cken, Germany \\
kissmann@cs.uni-saarland.de}
\and 
\'Alvaro Torralba 
\institute{Planning and Learning Group \\
Universidad Carlos III de Madrid, Spain \\
atorralb@inf.uc3m.es}
}
\begin{document}
\maketitle

\begin{abstract}
  For the exploration of large state spaces, symbolic search using
  binary decision diagrams (BDDs) can save huge amounts of memory and
  computation time. State sets are represented and modified by
  accessing and manipulating their characteristic functions. BDD
  partitioning is used to compute the image as the disjunction of
  smaller subimages.

  In this paper, we propose a novel BDD partitioning option. The
  partitioning is lexicographical in the binary representation of the
  states contained in the set that is represented by a BDD and uniform
  with respect to the number of states represented. The motivation of
  controlling the state set sizes in the partitioning is to eventually
  bridge the gap between explicit and symbolic search.

  Let $n$ be the size of the binary state vector. We propose an $O(n)$
  ranking and unranking scheme that supports negated edges and
  operates on top of precomputed satcount values. For the uniform
  split of a BDD, we then use unranking to provide paths along which
  we partition the BDDs. In a shared BDD representation the efforts
  are $O(n)$. The algorithms are fully integrated in the CUDD library
  and evaluated in strongly solving general game playing benchmarks.
\end{abstract}

\lstset{language=[GNU]C++,basicstyle=\small,numbers=left,numberstyle=\small}
\section{Introduction}

In this paper we are concerned with the space-efficient traversal of
state spaces with explicit-state search frontiers that are too large
to be kept in main memory. One of the options is a symbolic
representation of state sets in form of characteristic functions,
which are manipulated in the exploration.

Following this approach, \emph{binary decision diagrams} (BDDs)
\cite{bdd86} have been shown to be an efficient data structure for
representing and exploring large state sets in Model
Checking~\cite{bddRelProd} and Planning~\cite{Cimatti},
For some domains they can save a tremendous amount of memory (wrt.\ an
explicit representation). BDDs are also used as space-efficient
construction \emph{pattern database} search
heuristics~\cite{CompressionBDDs}.

\emph{General game playing} (GGP) goes in a similar direction as
automated action planning, but it can be seen to be even more general.
In general game playing the agent, i.e., the player, has to handle any
game that can be described by the used description language without
intervention of a human, and also without the programmer knowing what
game will be played -- it has to come up with a good strategy on its
own. The most used description mechanism is the \emph{game description
  language} GDL~\cite{gdl_specification}. A BDD classification
algorithm for (strongly) solving general single- and two-player games
has been proposed in~\cite{kissmann:layer-abstraction}.

A \emph{rank} is a number uniquely representing a state and the
inverse process, called unranking, reconstructs the state given its
rank. The approach advocated in this paper builds on top of findings
of~\cite{rank}, who illustrated that ranking and unranking of states
in a state set represented as a BDD is available in time linear to the
length of the state vector (in binary representation). In other words,
BDD ranking aims at the symbolic equivalent of constructing a perfect
hash function in explicit-state space search~\cite{PHF}. For the
construction of the perfect hash function, the underlying state set to
be hashed is generated in advance in form of a BDD. This is plausible
when computing \emph{strong solutions} to problems, i.e., the
game-theoretical value for each reachable state. Applications are,
e.g., endgame databases, or planning tasks where the problem to be
solved is harder than computing the reachability set.

Perfect hash functions to efficiently rank and unrank states have been
shown to be very successful in traversing single-player games, such as
Rubik's Cube or the Pancake Problem~\cite{TwoBitsBFS}, or two-player
games like Awari~\cite{AwariSolved}. They are also used for creating
pattern databases~\cite{BreyerK10a}. The problem of the construction
of perfect hash functions for algorithms like \emph{two-bit
  breadth-first search} is that they are problem-dependent.
 
\emph{BDD partitioning} approaches have been
proposed to address the so-called state-explosion
problem, which refers to the observation that the size of a state
space of a system tends to grow exponentially in the number of its
variables. In this paper we indicate that \emph{lex-partitioning},
short for \emph{lexicographical partitioning}, can advance symbolic
state space search.

As our approach refines the image operation, it applies to most BDD
exploration approaches in AI and beyond. We start with an introduction
to BDDs, to symbolic search and to computing strong solutions in
general games (Section \ref{sec:bdd}), as well as to the basic idea of
partitioning BDDs (Section \ref{sec:part}). Next, we turn to ranking
and unranking (Section \ref{sec:rank}), extended to the setting where
BDDs can have negated edges. Then, we consider the partitioning of a
BDD into sub-BDDs of an equal number of satisfying assignments
(Section \ref{sec:split}). Finally, we provide results in solving
general games (Section \ref{sec:eval}), discuss further implications
and conclude (Section \ref{sec:conclusion}).

\section{Binary Decision Diagrams for Strongly Solving Games}
\label{sec:bdd}

BDDs are a memory-efficient data structure used to represent Boolean
functions as well as to perform set-based search. In short, a BDD is a
directed acyclic graph with one root and two terminal nodes, the $0$-
and the $1$-sink. Each internal node corresponds to a binary variable
and has two successors, one (along the Else-edge) representing that
the current variable is false (0) and the other (along the Then-edge)
representing that it is true (1). For any assignment of the variables
derived from a path from the root to the $1$-sink the represented
function will be evaluated to~1.

Bryant \cite{bdd85} proposed a fixed variable ordering, for which he
also provided two reduction rules (eliminating nodes with the same
Then- and Else-successor and merging two nodes representing the same
variable that share the same Then-successor as well as the same
Else-successor). These BDDs are called reduced ordered binary decision
diagrams (ROBDDs). Whenever we mention BDDs in this paper, we actually
refer to ROBDDs. We also assume that the variable ordering is the same
for all the BDDs and has been optimized prior to the search.


BDDs have been shown to be very effective in the verification of hard-
and software systems, where BDD traversal is referred to as
\emph{symbolic model checking}~\cite{bddRelProd}. Adopting terminology
to state space search, we are interested in the \emph{image} of a
state set $S$ with respect to a transition relation $\mbox{\em
  Trans}$. The result is a characteristic function of all states
reachable from the states in $S$ in one step.

The image {\em Succ} of the state set $S$ is computed as $\mbox{\em
  Succ}(x') = \exists x \ (\mbox{\em Trans}(x,x')\ \wedge\ S(x))$. The
\emph{preimage}, which determines all predecessors of the state set
$S$, is computed as $\mbox{\em Pre}(x) = \exists x' \ (\mbox{\em
  Trans}(x,x')\ \wedge\ S(x'))$.


Using the image operator implementing a symbolic breadth-first search
(BFS) is straight-forward. All we need to do is to repeatedly apply
the image operator to the set of states reachable from the initial
state found so far. The search ends when a fix-point is reached, i.e.,
when no new successor states can be found. We store the set of all
reachable states as one BDD, so that, due to the structure of a BDD,
this does not contain any duplicates of states.

In general game playing (GGP) we are concerned with the problem of
automatically playing a game that the player probably has never seen
before, which is very similar to the action planning. There are
several differences, the first and foremost of course being that in
GGP we are not restricted to only one player, but rather an arbitrary
number of participants is supported. While in classical action
planning the goal is to find a plan, i.e., a sequence of actions
transforming the initial state to a goal state, as short as possible,
in GGP each terminal state has a specific outcome for each
participating player and the goal is to maximize the own outcome.

Allis \cite{allisPhD} proposed three kinds of \emph{solutions} for
two-player zero-sum games. In practice a \emph{weak solution} is often
enough, as it allows the game to be played \emph{optimally} in the
sense that the player following the solution will never achieve an
outcome worse than what was predicted for the initial state,
independent of the moves the opponent chooses. A problem arises only
when not following the solution at some step; in that case it might be
that the game-theoretic value as well as the best move to take are not
known for the successor state. In GGP this problem might arise when we
first played following some heuristic that told us what to do and
finished the calculation of the solution only afterward. That is why
we chose to calculate \emph{strong solutions}, which corresponds to
finding the game-theoretic value for each reachable state, and thus to
be able to determine the best move to take for every state that might
ever be encountered.

For the case of single-player games we might also speak about weak and
strong solutions. In that case a weak solution corresponds to a plan
that lets us reach the best possible outcome from the initial state,
while a strong solution again tells us the best possible outcome for
each reachable state, so that we can continue playing optimally even
after a suboptimal move has been chosen.


In order strongly solve single- or two-player
games~\cite{kissmann:layer-abstraction}, we find all the reachable
states by performing symbolic BFS, but instead of storing all
reachable states in one BDD we store each layer separately. The
solving starts in the last reached layer and performs regression
search towards the initial state, which resides in layer 0. This final
layer contains only terminal states (otherwise the forward search
would have progressed further), which can be solved immediately by
calculating the conjunction with the BDDs representing the rewards for
the two players. Once this is done, the search continues in the
preceding layer, because the remainder of the layer is empty. If
another layer contains terminal states as well, those are solved in
the same manner before continuing with the remaining states of that
layer. The rewards are handled in a certain order (e.g., in the order
win--draw--loss for the currently active player in case of a zero-sum
game or in decreasing order in case of a single-player game). All the
solved states of the successor layer are loaded in this order and the
preimage is calculated, which results in those states of the current
layer that will achieve the same rewards and are thus solved.

\section{Partitioning}
\label{sec:part}

For several domains constructing a transition relation \emph{Trans}
prior to the search consumes huge amounts of the available
computational resources. Fortunately, it is not required to build
$\mbox{\em Trans}$ monolithically, i.e., as one big relation.

Provided a set of actions $\Actions$, we can partition $\mbox{\em
  Trans}$ into individual transition relations $\mbox{\em
  Trans}_{\action}$ for each action $\action \in \Actions$, s.t.\
$\mbox{\em Trans} = \bigvee_{\action \in \Actions} \mbox{\em
  Trans}_{\action}$. For such a \emph{disjunctive partitioning} of the
transition relation the image now reads as
\begin{displaymath}
  \mbox{\em Succ}(x') = \exists x \ \left(\bigvee_{\action \in
      \Actions} \mbox{\em Trans}_{\action}(x,x') \ \wedge\ S(x)\right) =
  \bigvee_{\action \in \Actions} \left(\exists x \ (\mbox{\em
      Trans}_{\action}(x,x') \ \wedge\ S(x))\right).
\end{displaymath}
This image computation applies disjunctive splits for the different
actions to be applied and can accelerate BDD exploration compared to a
monolithical representation. One reason is that the relational product
for computing an image results in many intermediate BDDs and reveals
an NP hard problem (3-SAT can be reduced to an image operation
\cite{thesiskenneth}). The execution sequence of the disjunction has
an effect on the overall running time. In this case, we organize the
partitioned image in form of a binary tree, trying to have
intermediate BDDs of similar size.

A partitioning of $S$ into $k$ disjoint sets $S_1,\ldots,S_k$ ($S_i
\cap S_j = \emptyset$ for $i\neq j$) can lead to further simplified
sub-images, so that we have
\begin{displaymath}
  \mbox{\em Succ}(x') = \bigvee_{1\le l \le k} \bigvee_{\action \in \Actions}
  \left(\exists x \ (\mbox{\em Trans}_{\action}(x,x') \ \wedge\ S_l(x))
  \right).
\end{displaymath} 
Our partitioning method refines the following notion of a partitioned
BDD.

\begin{definition} [Partitioned BDD \cite{569909}]
  Given a Boolean function $f:\{0,1\}^n \rightarrow \{0,1\}$, defined
  over $n$ inputs $X_n = \{x_1,\ldots,x_n\}$, the \emph{partitioned
    BDD} representation is a set of $k$ function pairs $(w_1,f_1),
  \ldots, (w_k,f_{k})$, where $w_i,f_i: \{0,1\}^n \rightarrow \{0,1\}$
  are also defined over $X_n$ and satisfy the following four
  conditions.
  \begin{enumerate}
  \item $w_i$ and $f_i$ are represented as BDDs respecting the same
    variable ordering as $f$, for $1\le i \le k$.
  \item $w_1 \vee \ldots \vee w_k = 1$.
  \item $w_i \wedge w_j = 0$, for all $i \neq j$.
  \item $f_i = w_i \wedge f$, for $1\le i \le k$.
  \end{enumerate}
\end{definition}

We refer to the lexicographical ordering of bitvectors by using the
subindex \emph{lex}: for $a,b \in \{0,1\}^n$ we have $a <_{lex} b$ if
there is an $i \in \{1,\ldots,n\}$ such that $a_i < b_i$ and for all
$j \in \{1,\ldots,i-1\}$ we have $a_j = b_j$. Moreover, $a \le_{lex}
b$ iff $a <_{lex} b$ or $a= b$.

\begin{definition} [Lex-Partitioned BDD]
  Given a Boolean function $f:\{0,1\}^n \rightarrow \{0,1\}$, defined
  over $n$ inputs $X_n = \{x_1,\ldots,x_n\}$, the
  \emph{lex-partitioned BDD} representation of $f$ is a set of $k$
  assignments $a_1\ldots,a_{k} \in \{0,1\}^n$ and $k$ functions $f_1,
  \ldots, f_{k}:\ \{0,1\}^n \rightarrow \{0,1\}$ that are also defined
  over $X_n$ and satisfy the following conditions.
  \begin{enumerate}
  \item $f_i$ are represented as BDDs respecting the same variable
    ordering as $f$, for $1\le i \le k$.
  \item $a_k = (1,\ldots,1)$ and, for all $i < k$, we have $a_i
    <_{lex} a_{i+1}$.
  \item $f_1 \vee \ldots \vee f_k = f$.
  \item $f_i \wedge f_j = 0$ for all $i \neq j$.
  \item $f_1 = f \wedge \bigvee_{a \le_{lex} a_{1}} a$ and $f_i = f
    \wedge \bigvee_{a_{i-1} <_{lex} a \le_{lex} a_{i}} a$, for all $1
    < i \le k$.
  \end{enumerate}
\end{definition}

Using coefficients $w_1 = \bigvee_{a \le_{lex} a_{1}} a$ and $w_i =
\bigvee_{a_{i-1} <_{lex} < a \le_{lex} a_{i}} a$ for $1 < i \le k $
the definition specializes the one of partitioned BDDs. The advantage
is that by the lexicographical ordering we obtain more control over
the evolution of BDDs resulting from a split.

\section{Ranking and Unranking}
\label{sec:rank}

Linear-time ranking and unranking functions with BDDs have been given
in~\cite{rank}. \emph{Ranking} is a minimal perfect hash function from
the set of satisfying assignments to the position of it in the
lexicographical ordering of all satisfying assignments.
\emph{Unranking} is the inverse operation to ranking.

\begin{definition} [Ranking and Unranking]
  The \emph{rank} of an assignment $a \in \{0,1\}^n$ is the position
  in the lexicographical ordering of all satisfying assignments of the
  Boolean function $f$, while the \emph{unranking} of a number $r$ in
  $\{0,\ldots,C_f-1\}$ is its inverse, with $C_f$ being the total
  number of satisfying assignments of $f$.
\end{definition}

We have implemented the pseudo-code algorithms for the CUDD library
\cite{cudd}. The proposal in \cite{rank} does not support negated
edges. Negated edges, however, are crucial, since otherwise function
complementation is not a constant time operation, at least for a BDD
in a shared representation~\cite{Minato}.

\begin{definition} [Edge Complementation, Satcount, Conversion]
  The \emph{index} of a BDD node is its unique position in the shared
  representation. For the ease of notation we take the negation of the
  node index to represent \emph{edge complementation}, i.e., $-n$ is
  the negated and $|n|$ is the regular node index. The function
  $\mbox{\em sign}(n)$ returns $-1$ if the edge is complemented and
  $1$ if it is not, $\mbox{\em variable}(n)$ returns the variable
  associated with $n$ and $\mbox{\em level}(n)$ its position in the
  variable ordering. Moreover, we assume the 1-sink to have the node
  index $1$ and the 0-sink to have the node index $-1$. Let $C_f =
  |\{a \in \{0,1\}^n \mid f(a) = 1\}|$ denote the number of satisfying
  assignments (\emph{satcount}) of $f$. With $\mbox{\em bin}$ (and
  $\mbox{\em invbin}$) we denote the \emph{conversion} of the binary
  value of a bitvector (and the inverse operation).
\end{definition}

For ranking and unranking the satcount values are precomputed for
every essential subfunction and stored in the unique table for the
shared BDD. This table is used by two functions: $\mbox{\em
  insert}(n,v)$ sets a value $v$ for a node $n$ in the unique table
and $\mbox{\em lookup}(n)$ retrieves it. Memory is allocated if a node
is new.

As BDDs are reduced, not all variables on a path are present, but need
to be accounted for in the satcount procedure.
Figure~\ref{fig:satcount} shows the pseudo-code of the function that
does not only compute the values but also stores all the intermediate
results and follows the proposal of~\cite{bdd86}. We see that the time
(and space) complexity is $O(|G_f|)$, where $|G_f|$ is the number of
nodes of the BDD $G_f$ representing $f$.

\begin{figure}[t]
  \centering
  \begin{minipage}{10cm}
    \begin{lstlisting}
precomputeSatCount()
  i = level(root);
  satcount = 2^i * satCountAux(root);
\end{lstlisting}
\begin{lstlisting}
satCountAux(n)
  if (n == 1-sink()) return 1;
  if (n == 0-sink()) return 0;
  if (res = lookup(n)) return res;
  t = sign(n) * Then(|n|); e = sign(n) * Else(|n|);
  i = level(|n|); j = level(t); k = level(e);
  satcount = (2^(j-i-1)) * satCountAux(t) +
             (2^(k-i-1)) * satCountAux(e);
  insert(n,satcount);
  return satcount;
\end{lstlisting}
\end{minipage}
\caption{Satisfiability Counting with Negated Edges.}
\label{fig:satcount}
\end{figure}   

With negation on edges there are subtle problems to be resolved for
storing the satcount values. While the number of satisfiable paths for
a node might fit into a computer word this is not necessarily true for
the negated subfunction. Therefore, we allow up to two satisfiability
values to be stored together with a node: one wrt.\ reaching it on a
negated edge, and the other one wrt.\ reaching it on a non-negated
edge. In contrast to standard satisfiability count implementations (as
in CUDD) this way we ensure that only satcount values of at most $c$
are stored, where $c$ is the satcount value of the root node. E.g., in
ConnectFour $7 \times 6$ with $85$ binary variables (yielding $2^{85}$
possible values), long integers are sufficient to store intermediate
satcount values, which are all smaller than the satcount value of the
entire reachable set ($\numprint{4531985219092}$).

Figures~\ref{fig:ranking} and~\ref{fig:unranking} extend the proposal
of \cite{rank} and show the ranking and unranking functions and thus
realize an invertible minimal perfect hash function for $f$ mapping an
assignment $s \in \{0,1\}^n$ to a value $r \in \{0,\ldots,C_f-1\}$.
The procedures determine the rank given a satisfying assignment and
vice versa. They access the satcount values on the Else-successor of
each node (adding for the ranking and subtracting for the unranking).
Missing nodes (due to BDD reduction) have to be accounted for by their
binary representation, i.e., gaps of $l$ missing nodes are accounted
for $2^l$. Edge complementation changes the sign of the node $n$ and
is progressed to evaluation of the sinks. While the ranking procedure
is recursive the unranking procedure is not. Both procedures track the
gap imposed by the distance in the levels of the current and the
successor node

\begin{figure}[Htbht]
\centering
\begin{minipage}{10cm}
\begin{lstlisting}
rank(s)
  i = level(root);
  d = bin(s[0..i-1]);
  return d*satCount(root) + rankAux(root,s) - 1;
\end{lstlisting}
\begin{lstlisting}
rankAux(n,s)
  if (|n| == 1) return 1;
  t = sign(n) * Then(|n|); e = sign(n) * Else(|n|);
  i = level(|n|); j = level(e); k = level(t);
  if (s[i] == 0)
    return bin(s[i+1..j-1]) * satCount(e) + rankAux(e,s);
  else
    return 2^(j-i-1) * satCount(e) +
           bin(s[i+1..k-1]) * satCount(t) + rankAux(t,s);
\end{lstlisting}
\end{minipage}
\caption{Ranking with Negated Edges.}
\label{fig:ranking}
\end{figure} 

\begin{figure}[Hbht]
\centering
\begin{minipage}{10cm}
\begin{lstlisting}
unrank(r)
  i = level(root); d = r / satCount(root);
  s[0..i-1] = invbin(d);
  n = root;
  while (|n| > 1)
    r = r % satCount(n);
    t = sign(n) * Then(|n|); e = sign(n) * Else(|n|);
    j = level(e); k = level(t);
    if (r < (2^(j-i-1) * satCount(e)))
      s[i] = 0;
      d = r / satCount(e);
      s[i+1..j-1] = invbin(d);
      n = e; i = j;
    else
      s[i] = 1;
      r = r - (2^(j-i-1) * satCount(e));
      d = r / satCount(t);
      s[i+1..k-1] = invbin(d);
      n = t; i = k;
\end{lstlisting}
\end{minipage}
\caption{Unranking with Negated Edges.}
\label{fig:unranking}
\end{figure} 

Once the satcount values have been precomputed, both functions require
linear time $O(n)$, where $n$ is the number of variables in the
function represented in the BDD. Dietzfelbinger and Edelkamp provide
invariances showing that the procedures work correctly~\cite{rank}.

\section{Splitting}
\label{sec:split}

Given the BDD $G_f$ and any assignment $s \in \{0,1\}^n$, the
\emph{split} function computes the BDDs $G_g$ and $G_h$ with the
satisfying sets $S_g = S_f \cap \{b \in \{0,1\}^n \mid b \le_{lex} s \}$ and
$S_h = S_f \cap \{b \in \{0,1\}^n \mid b >_{lex} s \}$. If we choose the
assignment as the result of unranking $\lfloor C_f / 2 \rfloor$ we get
$C_g =\lfloor C_f / 2 \rfloor$ and $C_h= \lceil C_f / 2 \rceil$.

\begin{figure}
\centering
\begin{minipage}{10cm}
\begin{lstlisting}
pair split(s)
  return splitAux(root,0,s);

pair splitAux(n, lev, s)
  z = 0-sink();
  if (lev < level(|n|))
    (s1,s2) = splitAux(n,lev+1,s);
    if(s[lev])
      left = new node(var(lev),n,s1);
      right = new node(var(lev),z,s2);
    else
      left = new node(var(lev),s1,z);
      right = new node (var(lev),s2,n);
  else if (n == 1 or n == 0)
    left = n; right = z;
  else
    t = sign(n) * Then(|n|); e = sign(n) * Else(|n|);
    if(s[lev])
      (t1,t2) = splitAux(t,lev+1,s);
      left  = new node(variable(|n|),t1,e);
      right = new node(variable(|n|),t2,z);
    else
      (e1,e2) = splitAux(e,lev+1,s);
      left  = new node(variable(|n|),z,e1);
      right = new node(variable(|n|),t,e2);
  return (left, right);
\end{lstlisting}
\end{minipage}
\caption{Splitting with Edge Complements.}
\label{fig:spliting}
\end{figure} 

Figure \ref{fig:spliting} shows the pseudo-code of the recursive split
algorithm. The input is the state vector in form of an assignment along
which the BDD should be split. The result consists of two BDDs: the
left BDD represents all the assignments lexicographically smaller or
equal than the selected assignment $a$ and the right BDD all the
others. The algorithm traverses the path imposed by the input vector
$s$ bottom-up. Whenever needed, it allocates new nodes. If a node
already exists, no allocation takes place. Depending on the truth
value of the bitvector position $lev$ currently processed, we swap the
attachment of sub-BDDs.


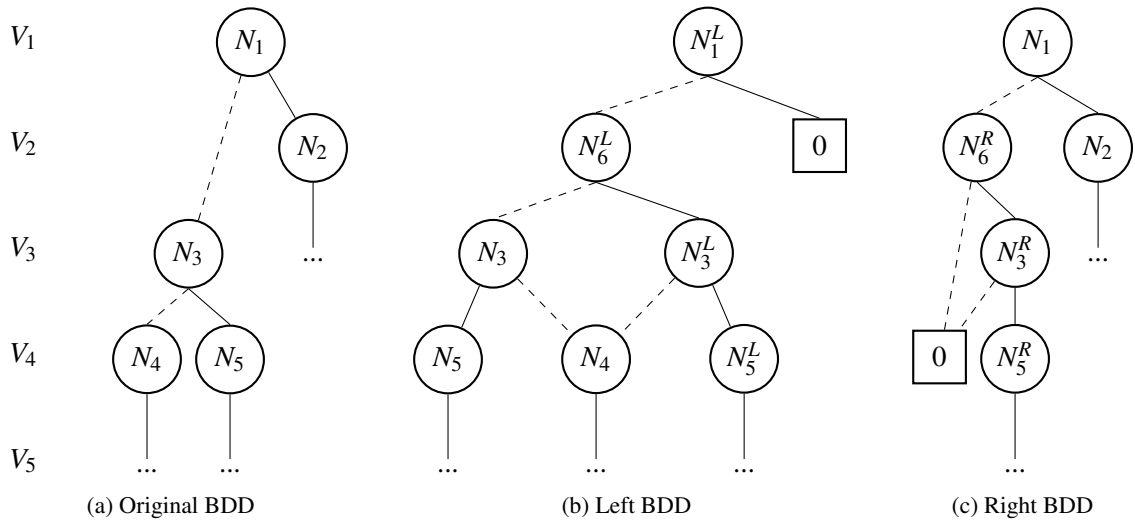
\begin{figure}
  \centering
\subfloat[Original BDD]{
\begin{tikzpicture}
\tikzset{edge from parent/.style=
{draw=none}}
\tikzset{level distance=40pt}
\tikzset{sibling distance=5pt}
\Tree [.\node{$V_1$};
  [.\node{$V_2$};
    [.\node{$V_3$};
      [.\node{$V_4$}; 
        [.\node{$V_5$}; 
     ]
     ]]]]
\end{tikzpicture}
\qquad
\begin{tikzpicture}
\tikzset{level distance=40pt}
\tikzset{sibling distance=5pt}
\tikzstyle{node}=[circle,draw,thick, inner sep=0pt,minimum size=9mm]
\tikzstyle{sink}=[draw, thick, inner sep=0pt,minimum size=7mm]
  \Tree 
      [.\node[node](n1){$N_{1}$}; \edge [color=white];
       [\edge [draw=none];
          [.\node[node, level distance=80pt](n3){$N_{3}$}; 
           \edge[dashed]; [.\node[node](n4){$N_{4}$}; ... ]
            [.\node[node](n5){$N_{5}$}; ... ]
          ]
        ]
        \edge [draw=none]; [.\node[node](n2){$N_{2}$}; ... ]
 ]
    \draw [dashed] (n1) -- (n3);
    \draw (n1) -- (n2);
\end{tikzpicture}}
\qquad
\subfloat[Left BDD]{
  \begin{tikzpicture}
\tikzstyle{node}=[circle,draw,thick, inner sep=0pt,minimum size=9mm]
\tikzstyle{sink}=[draw, thick, inner sep=0pt,minimum size=7mm]
\tikzset{level distance=40pt}
\tikzset{sibling distance=5pt}
  \Tree 
      [.\node[node](ln1){$N_1^L$};
       \edge[dashed]; [.\node[node](ln6){$N_6^L$};
         \edge[dashed]; [.\node[node](ln3){$N_3$}; 
            \edge [draw=none]; [.\node[node](ln5){$N_{5}$}; ... ]
            \edge [draw=none]; [.\node [missing]{};]
            \edge [draw=none]; [.\node [missing]{};]
          ]
            \edge [draw=none]; [.\node [missing]{}; 
            \edge [draw=none]; [.\node[node](ln4){$N_{4}$}; ... ]
            ]
          [.\node[node](ln32){$N_3^L$}; 
            \edge [draw=none]; [.\node [missing]{};]
            \edge [draw=none]; [.\node [missing]{};]
             \edge [draw=none]; [.\node[node](ln52){$N_{5}^L$}; ... ]
          ]
        ]
        [.\node[sink](n22){$0$};]
      ]
    \draw (ln3) -- (ln5);
    \draw [dashed] (ln3) -- (ln4);
    \draw [dashed] (ln32) -- (ln4);
    \draw (ln32) -- (ln52);
\end{tikzpicture}}
\qquad
\subfloat[Right BDD]{
\begin{tikzpicture}
\tikzstyle{node}=[circle,draw,thick, inner sep=0pt,minimum size=9mm]
\tikzstyle{sink}=[draw, thick, inner sep=0pt,minimum size=7mm]
\tikzset{level distance=40pt}
\tikzset{sibling distance=5pt}
  \Tree 
      [.\node[node](rn1){$N_{1}$};
        \edge[dashed]; [.\node[node](rn6){$N_{6}^R$}; 
          \edge[draw=none];[.\node[missing]{}; \edge[draw=none]; [.\node[sink](rn0){$0$};] ]
          [.\node[node](rn3){$N_{3}^R$}; 
            [.\node[node](rn5){$N_{5}^R$}; ... ]
          ]
        ]
        [.\node[node](rn2){$N_{2}$}; ... ]
      ]
    \draw[dashed] (rn6) -- (rn0);
    \draw[dashed] (rn3) -- (rn0);
\end{tikzpicture}
}
\caption {Example of the split algorithm. The original BDD is split
  into its left and right part with respect to the assignment
  $011\dots$ (Dashed lines represent Else-edges). }
\label{fig:splitexample}
\end{figure}

Figure \ref{fig:splitexample} shows how the algorithm works in a part
of the BDD. Each node in the path represented by the assignment $a =
011\dots$ is split into two, depending on the value of the assignment
for the associated variable.

If the assignment is~0, the recursion is made over the Else-edge. The
result of the recursion is set as the Else-edges of the left and right
part, respectively. The Then-edge points to the $0$-sink in the left
part and to the Then-successor of the original node in the right part.
In the example $N_{1}$ is divided into $N_{1}^L$ and $N_{1}^R$. All
the assignments with $V_1 = 1$ are considered in the right part, while
the others are split in the recursion. Symmetric rules apply in case
that the assignment is~1 (node $N_{3}$ in the example).

The base case corresponds to the constant node, returning the $1$-sink
for the left part and the $0$-sink for the right one, assigning $s$ to
the left part. Finally, if some node in the path is missing (due to
the elimination rule of nodes with the same Then- and Else-successor),
the algorithm still splits it into two following the same rules
described above. In the example, $N_6^L$ and $N_6^R$ are the result of
the split over a missing node in the original BDD with both edges
pointing to $N_3$.

\begin{theorem} [Time Complexity Split Function]
  In a shared BDD representation given the BDD $G_f$ and an assignment
  $a \in \{0,1\}^n$ the split function computes the BDDs $G_g$ and
  $G_h$ in $O(n)$ time.
  \begin{proof}
    As at most $n$ nodes are processed in post-order, the time
    complexity is immediate. Moreover, the BDDs that are constructed
    are reduced. All original nodes remain valid in the shared
    representation and each new node that is created in the bottom-up
    traversal is checked for applicability of the BDD reduction rules
    (by issuing a look-up in the unique table).
  \end{proof}
\end{theorem}
  
\begin{theorem} [Space Complexity Split Function]
  We have $|G_g| \le |G| + n$ and $|G_h| \le |G| + n$. In a shared BDD
  representation we also have $|G_g \cup G_h| \le |G| + 2n$.
  \begin{proof}
    As at most $2n$ nodes are created in the shared representation the
    second result $|G_g \cup G_h| \le |G| + 2n$ is immediate. For each
    individual function $G_g$ and $G_h$ we have constructed at most
    $n$ new nodes. If we extract a BDD from the shared representation
    we duplicate nodes from $G$ that are shared between the two
    structures.
  \end{proof}
\end{theorem}

By applying repeated splits we can
uniformly partition a BDD into $k$ parts in time $O(kn)$.






\section{Experimental Evaluation}
\label{sec:eval}

We have implemented our algorithms by extending the CUDD
package~\cite{cudd} to support the different satcount procedures,
ranking, unranking and various split options. The solver program is
written in Java 
and connected to Fabio Somenzi's BDD package CUDD with the
Java Native Interface (JNI).

We performed the experiments on one core of a 64-bit desktop computer
(model Intel(R) Xeon(R) CPU X3470 with $2.93$\,GHz) running Linux
(Ubuntu). This computer is equipped with $8$\,GB main memory and
$\numprint{8192}$\,KB cache. For the experiments there was no need to
use virtual memory. We compiled the CUDD package using the GNU
C\texttt{++} compiler (\verb|gcc| version 4.3 with option \verb|-O3|).

We conducted experiments in different games provided in the general
description language GDL~\cite{gdl_specification}. The selection of
games indicates the generality of the approach: seven single-player
games (8-Puzzle, Asteroids Parallel, Knight's Tour, Lightsout,
Lightsout 2, Peg Solitaire, and Tpeg) and five two-player games (Catch
a Mouse, CephalopodMicro, ConnectFour 5 $\times$ 5, NumberTicTacToe,
and TicTacToe) For the description of the games and their
implementations in GDL we refer to the commonly used GGP
server\footnote{http://ggpserver.general-game-playing.de}. We used a
timeout of one hour for every experiment.

For each game, the exploration is performed in two phases \cite
{kissmann:layer-abstraction}. First, a breadth-first search generates
the reachable states, organized in layers. Then, a backward
exploration classifies the states in each layer according to their
reward by computing the preimage of the classified sets of the next
layer. In case the explorations are completed, the games are strongly
solved, i.e., the game-theoretical value of each reachable state is
computed. This amounts to a combined forward and backward exploration
to compute the set of reachable states and to classify it into sets of
different game-theoretical values. The number of backward images is
usually greater than the number of forward images as different
classification sets have to be computed by calling the image operator.

We compare our partitioning method to others already implemented in
the CUDD library. There are different strategies, e.g., splitting for
balancing the number of states, for balancing the number of nodes, and
other disjunctive subset algorithms.

When balancing the number of states, it is possible to limit the
number of states in each BDD by splitting the original BDD in as many
parts as necessary (\emph{States}) or to split the BDD in a fixed
number of folds, all of them with the same number of states (\emph
{FoldStates}). To get partitions with the desired number of states we
make use of our lexicographic partitioning (\emph{Lex}). We do not report experiments with another state-selection strategy included in the CUDD library, given that there is not significant differences wrt. our lexicographic version. The main difference is that our version respect the lexicographic order, which may be an advantage when ranking/unranking states or when assigning states to different cores in a distributed version. 

In order to get partitions with balanced number of nodes we consider
also limiting the maximum number of nodes in each BDD (\emph{Nodes})
or splitting the BDD in a fixed number of parts with balanced
number of nodes (\emph{FoldNodes}).
The CUDD library includes several algorithms that allow splitting a
BDD according to the number of nodes:
\begin{itemize}
\item Shortest Path (\emph{SPath}): Procedure to subset the given BDD
  choosing the shortest paths (largest cubes) in the BDD.
\item Compress (\emph{Comp}): Finds a dense subset using several
  techniques in series. It is more expensive than other subsetting
  procedures, but often produces better results.
\end{itemize}

There are also other methods that allow for a disjunctive
decomposition in two parts (\emph{Disj}) according to different criteria:
\begin{itemize}
\item Iterative (\emph{Ite}): Iterated use of supersetting to obtain a
  factor of the given function. The two parts tend to be imbalanced.
\item Generation (\emph{Gen}): generalizes the decomposition based on
  the cofactors with respect to one variable.
\item Variable selection (\emph{Var}): Decomposes the BDD according to
  the value of a variable, chosen to minimize and balance the size of
  the resulting BDDs.
\end{itemize}

\begin{table}[hbtp] \centering 
\begin{tabular}{r|rrrrrrrrr}
 & FStates & States & FNodes & FNodes & Nodes & Nodes & Disj & Disj & Disj\\
 & Lex & Lex & Comp & SPath & Comp & SPath & Var & Ite & Gen\\
 & 8 & 100000 & 8 & 8 & 10000 & 10000 &  &  & \\
\hline8-puzzle & 1.00 & 1.00 & 1.00 & 1.00 & 1.00 & 1.00 & 1.00 & 1.00 & 1.00\\\hline
ConnectFour $5\times 5$ & 1.00 & 1.00 & 1.00 & 1.00 & 1.00 & 1.00 & 1.00 & 1.00 & 1.00\\\hline
Knights Tour & 1.00 & 1.00 & 1.00 & 1.00 & 1.00 & 1.00 & 1.00 & 1.00 & 1.00\\\hline
Lightsout & 1.00 & 1.00 & 1.00 & 1.00 & 1.00 & 1.00 & 1.00 & 1.00 & 1.00\\\hline
Peg & 0.84 & 0.76 & 0.79 & 0.82 & 0.53 & 0.56 & 0.79 & 0.79 & 0.71\\\hline
Asteroids Parallel & 1.04 & 0.11 & 0.99 & 0.99 & 0.10 & 0.10 & 0.99 & 0.99 & 0.99\\\hline
Lightsout 2 & 1.00 & 1.00 & 1.00 & 1.00 & 1.00 & 1.00 & 1.00 & 1.00 & 1.00\\\hline
TPeg & 0.89 & 0.80 & 0.88 & 0.89 & 0.14 & 0.15 & 0.85 & 0.89 & 0.81\\\hline
Catcha Mouse & 1.00 & 0.22 & 1.00 & 1.00 & 1.00 & 1.00 & 1.00 & 1.00 & 1.00\\\hline
Cephalopod Micro & 0.91 & 0.69 & 0.85 & 0.81 & 0.19 & 0.19 & 0.88 & 0.91 & 0.81\\\hline
Number TicTacToe & 1.00 & 1.00 & 1.00 & 1.00 & 1.00 & 1.00 & 1.00 & 1.00 & 1.00\\\hline
TicTacToe & 1.00 & 0.19 & 1.00 & 1.00 & 0.55 & 0.55 & 1.00 & 1.00 & 1.00\\\hline
\hline Total & 0.97 & 0.76 & 0.95 & 0.95 & 0.68 & 0.69 & 0.95 & 0.96 & 0.93\\\hline
\end{tabular}
\caption{Number of computed layers by each partitioning scheme wrt. not applying any partitioning}
\label{table:coverage}
\end{table}

Table~\ref{table:coverage} shows a comparison of the number of layers explored
using each partitioning before the timeout. In half of the domains all the partitioning versions are able to finish the exploration. In the other domains most partitions are close to the exploration without partitioning solving more than 90\% of the layers, except the three versions without a bounded number of partitions (\emph{States Lex 100000}, \emph{Nodes Shortest Path 10000} and \emph{Nodes Compress 10000}).
Fold States Lex 8 dominates the other partitioning methods in all the domains, being the only one able to compute more layers than the version without partitioning in one domain: Asteroids Parallel.

\begin{table}[hbtp] \centering 
\begin{tabular}{r|rrrrrrrrr}
 & FStates & States & FNodes & FNodes & Nodes & Nodes & Disj & Disj & Disj\\
 & Lex & Lex & Comp & SPath & Comp & SPath & Var & Ite & Gen\\
 & 8 & 100000 & 8 & 8 & 10000 & 10000 &  &  & \\
\hline8-puzzle & 1.17 & 1.04 & 1.77 & 1.43 & 1.25 & 1.07 & 1.91 & 1.11 & 2.15\\\hline
ConnectFour $5\times 5$ & 1.44 & 1.57 & 3.24 & 2.28 & 7.72 & 5.74 & 2.97 & 1.50 & 3.75\\\hline
Knights Tour & 1.37 & 1.40 & 2.66 & 1.92 & 5.45 & 4.24 & 2.90 & 1.45 & 3.05\\\hline
Lightsout & 1.21 & 1.23 & 1.56 & 1.32 & 2.63 & 2.20 & 1.57 & 1.26 & 2.47\\\hline
Peg & 1.03 & 1.06 & 1.04 & 1.05 & 1.13 & 1.08 & 1.05 & 1.03 & 1.03\\\hline
Asteroids Parallel & 0.60 & 0.35 & 0.95 & 0.97 & 0.18 & 0.19 & 0.50 & 0.66 & 0.76\\\hline
Lightsout 2 & 1.22 & 1.23 & 1.56 & 1.32 & 2.65 & 2.21 & 1.57 & 1.26 & 2.46\\\hline
TPeg & 1.02 & 1.04 & 1.02 & 0.98 & 1.10 & 1.08 & 1.04 & 1.01 & 1.03\\\hline
Catcha Mouse & 1.71 & 166.53 & 1.73 & 1.29 & 1.21 & 1.06 & 2.22 & 1.01 & 1.79\\\hline
Cephalopod Micro & 1.03 & 1.21 & 1.05 & 1.07 & 1.21 & 1.17 & 1.05 & 1.03 & 1.10\\\hline
Number TicTacToe & 1.18 & 1.20 & 1.71 & 1.89 & 2.38 & 2.55 & 1.55 & 1.42 & 2.16\\\hline
TicTacToe & 1.84 & 3.30 & 1.88 & 1.65 & 2.71 & 2.55 & 1.73 & 1.34 & 2.96\\\hline
\hline Total & 0.99 & 1.23 & 1.12 & 1.09 & 1.15 & 1.08 & 1.01 & 0.97 & 1.19\\\hline
\end{tabular}
\caption{Relative time spent in solving each game for each partitioning scheme wrt. not applying any partitioning}
\label{table:time}
\end{table}

\begin{table}[hbtp] \centering 
\begin{tabular}{r|rrrrrrrrr}
 & FStates & States & FNodes & FNodes & Nodes & Nodes & Disj & Disj & Disj\\
 & Lex & Lex & Comp & SPath & Comp & SPath & Var & Ite & Gen\\
 & 8 & 100000 & 8 & 8 & 10000 & 10000 &  &  & \\
\hline8-puzzle & 0.36 & 1.00 & 0.83 & 0.80 & 0.98 & 0.86 & 0.79 & 0.72 & 0.82\\\hline
ConnectFour $5\times 5$ & 0.23 & 0.15 & 0.85 & 0.57 & 0.84 & 0.74 & 0.62 & 0.60 & 0.87\\\hline
Knights Tour & 0.20 & 0.22 & 0.56 & 0.56 & 0.56 & 0.56 & 0.58 & 1.00 & 0.92\\\hline
Lightsout & 0.51 & 0.34 & 0.76 & 0.70 & 0.78 & 0.73 & 0.78 & 0.78 & 0.82\\\hline
Peg & 0.25 & 0.06 & 0.11 & 0.58 & 0.02 & 0.55 & 0.68 & 0.65 & 0.68\\\hline
Asteroids Parallel & 0.28 & 0.01 & 0.29 & 0.19 & 0.00 & 0.01 & 0.55 & 0.51 & 1.09\\\hline
Lightsout 2 & 0.51 & 0.34 & 0.76 & 0.70 & 0.78 & 0.73 & 0.78 & 0.78 & 0.82\\\hline
TPeg & 0.23 & 0.05 & 0.62 & 0.52 & 0.02 & 0.58 & 0.67 & 0.61 & 0.70\\\hline
Catcha Mouse & 1.00 & 1.00 & 1.00 & 1.00 & 1.00 & 1.00 & 1.00 & 1.00 & 1.00\\\hline
Cephalopod Micro & 0.22 & 0.01 & 0.90 & 0.61 & 0.01 & 0.42 & 0.61 & 0.73 & 0.65\\\hline
Number TicTacToe & 0.27 & 0.27 & 1.01 & 0.96 & 0.10 & 0.98 & 0.65 & 0.79 & 0.86\\\hline
TicTacToe & 0.41 & 0.00 & 0.10 & 0.96 & 0.04 & 0.84 & 0.83 & 0.93 & 1.31\\\hline
\hline Total & 0.29 & 0.04 & 0.40 & 0.46 & 0.07 & 0.35 & 0.63 & 0.64 & 0.96\\\hline
\end{tabular}
\caption{Relative maximum number of nodes per image for each partitioning scheme wrt. not applying any partitioning}
\label{table:memory}
\end{table}

Tables~\ref {table:time} and \ref{table:memory} show the relative time
spent in solving each game and the maximum number of nodes per image,
respectively, for each partitioning configuration wrt.\ the ``None''
partitioning version. For those games where some partitioning strategy did
not finish due to a timeout, only layers solved by the algorithm
and the ''None'' partitioning version were taken into account.

In distributed or external memory settings the maximum number of nodes involved in a single image determines the memory needed. 
Thus, the results show that BDD partitioning can help to solve problems by
reducing the memory requirements, at the cost of increasing the time
spent. Versions splitting the BDDs in an unbounded number of parts
achieve good memory reductions but they do not scale well, both when
balancing the BDDs according to the number of states or according to
the number of nodes. On the other hand, versions applying a
partitioning in 2 or 8 folds do not impose a large overhead, being
around 50 percent slower than the no-partitioning version.

When comparing the way to select subsets, in general the lexicographic
partitioning allows great reductions in the necessary memory while not
producing a large overhead in time. It gets the best coverage results
among all partitioning methods, being the only one able to improve the coverage
of the non-partitioning version in one domain.
Furthermore, of all the strategies with a fixed number of partitions, 
it is the one achieving largest memory savings and the second best in overall time.

\begin{figure}[hbtp]
\centering
\subfloat[8 Puzzle]{\includegraphics[width=7.6cm]{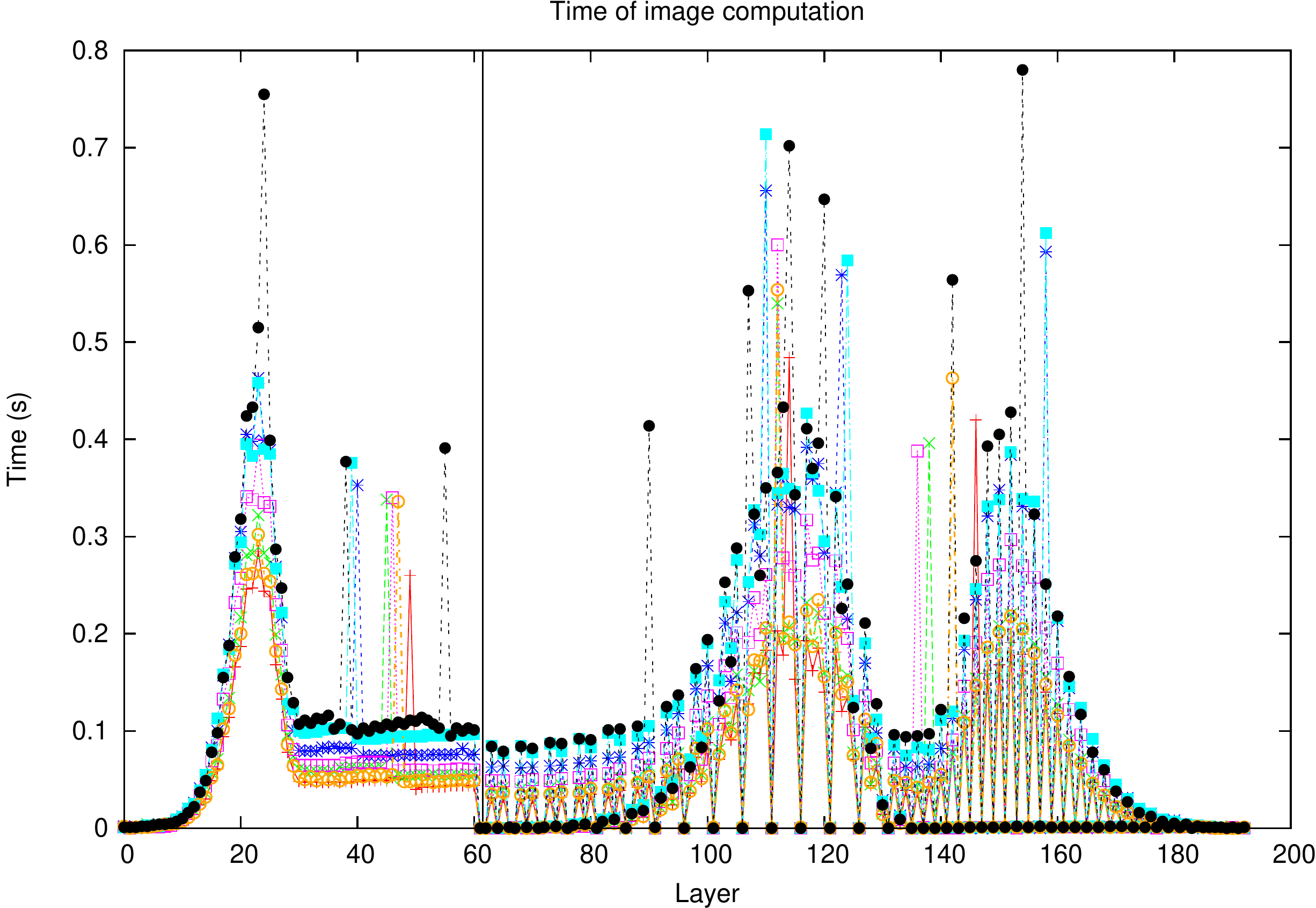}}
\qquad
\subfloat[Connect Four $5\times 5$]{\includegraphics[width=7.6cm]{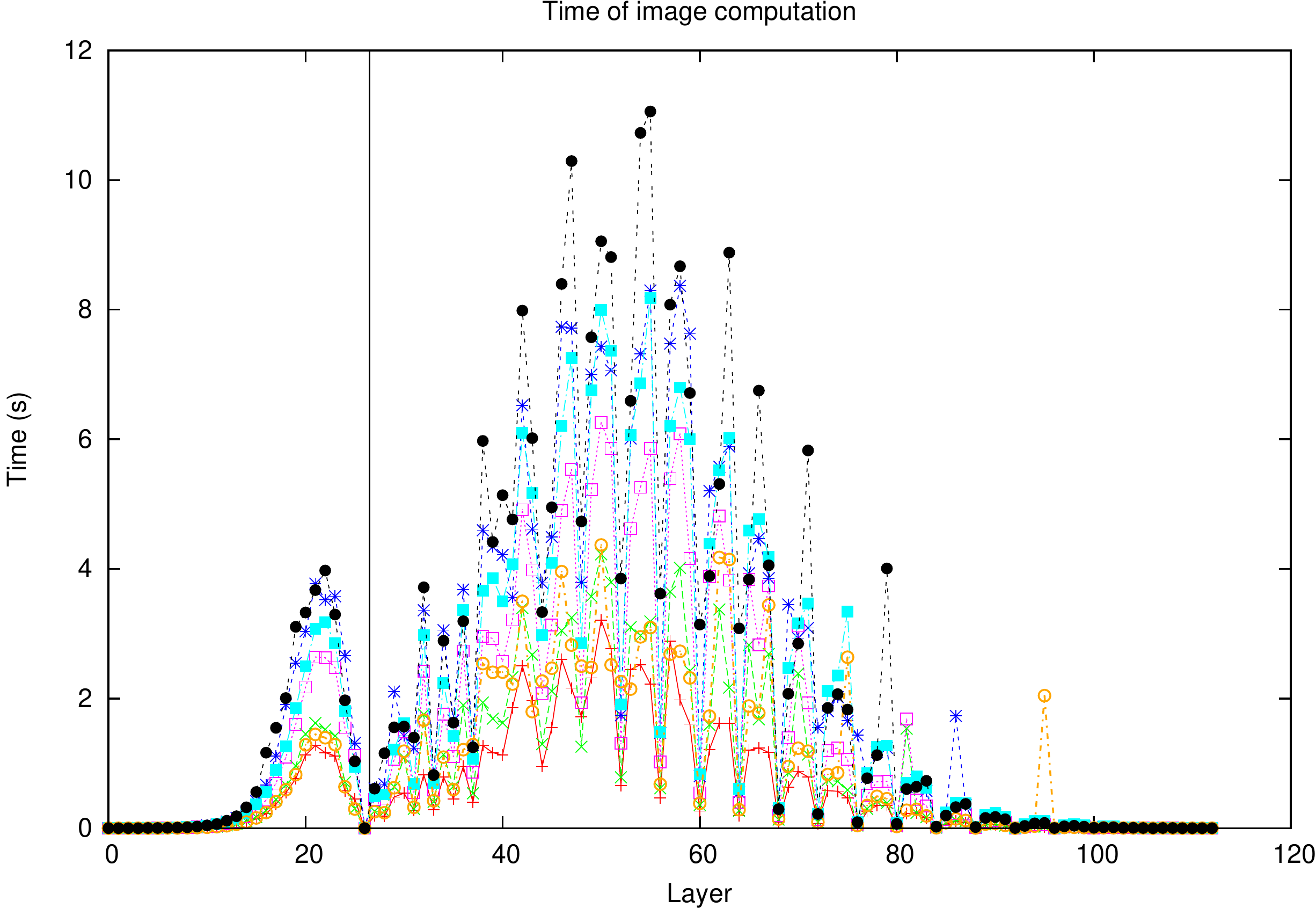}}
\\
\subfloat[Knights Tour]{\includegraphics[width=7.6cm]{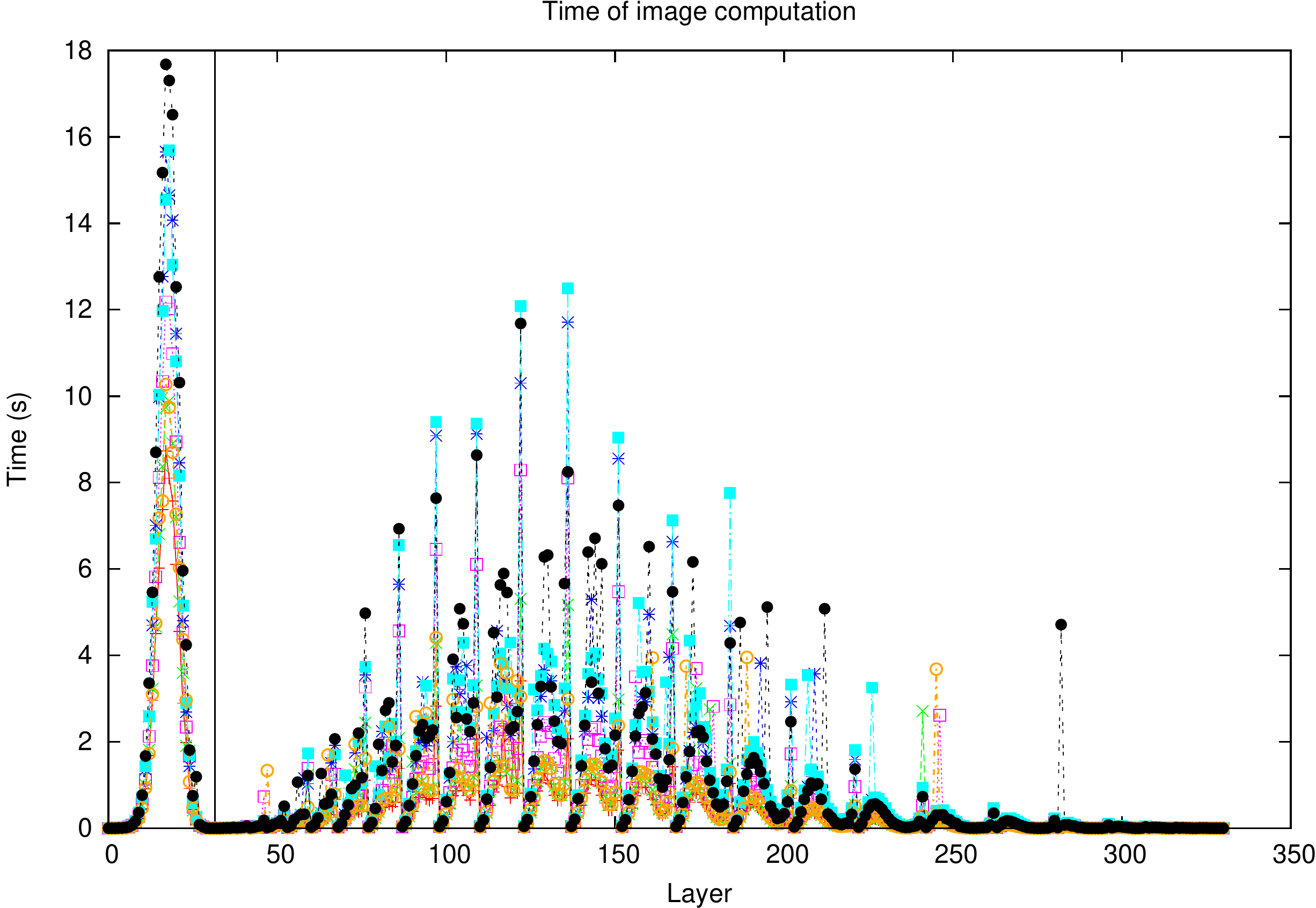}}
\qquad
\subfloat[Lightsout]{\includegraphics[width=7.6cm]{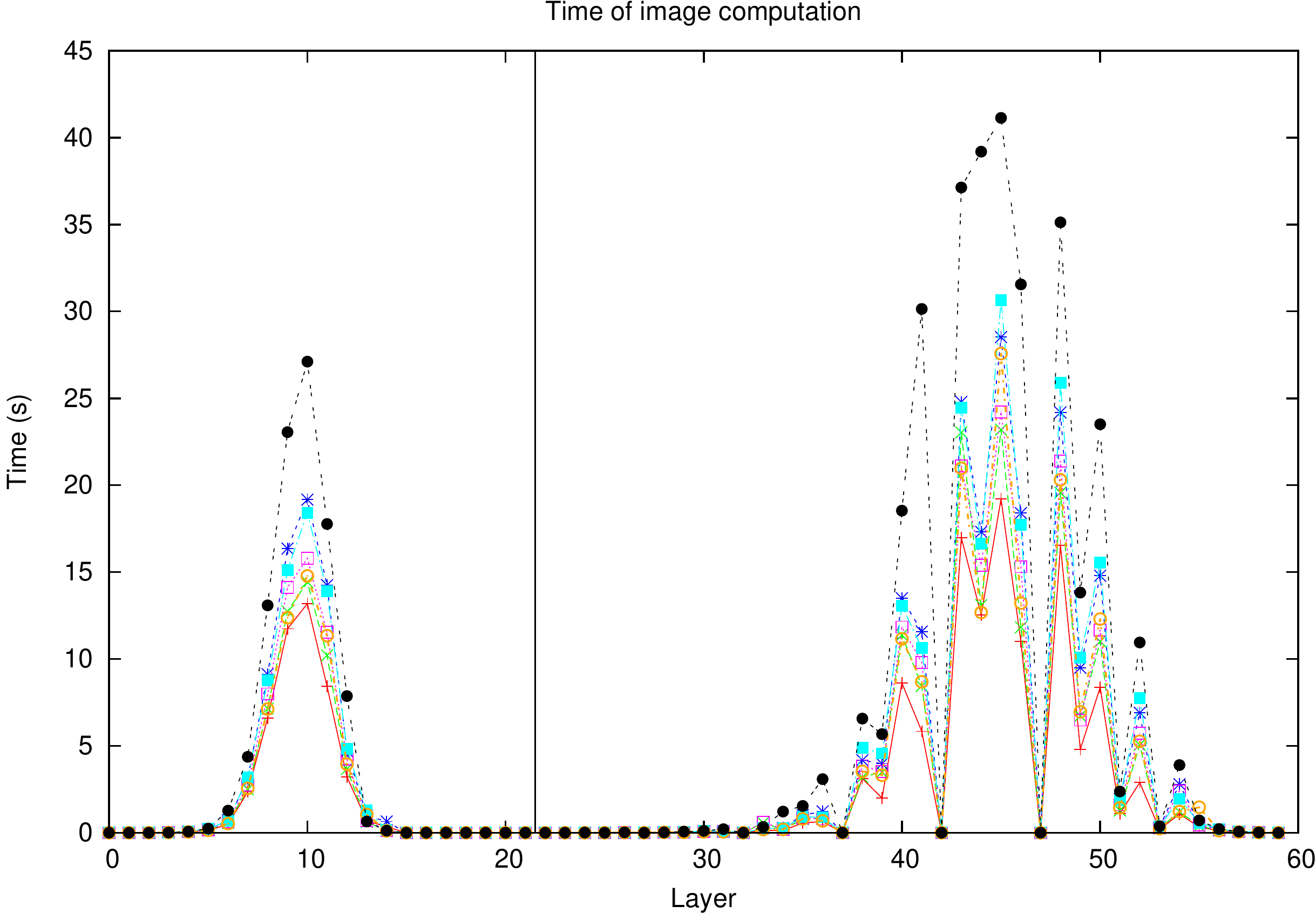}}
\\
\subfloat[TPeg]{\includegraphics[width=7.6cm]{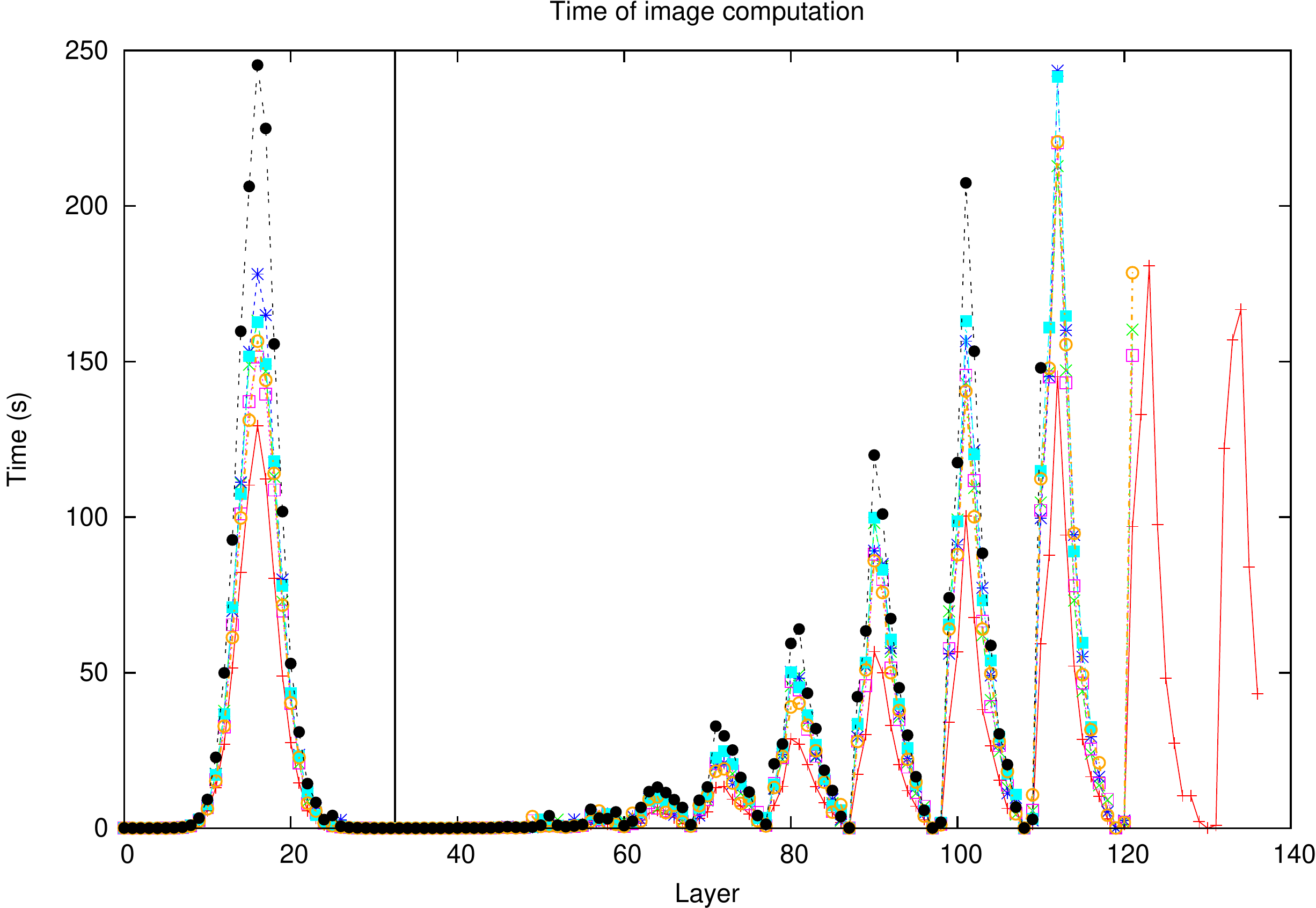}}
\qquad
\subfloat[Asteroids Parallel]{\includegraphics[width=7.6cm]{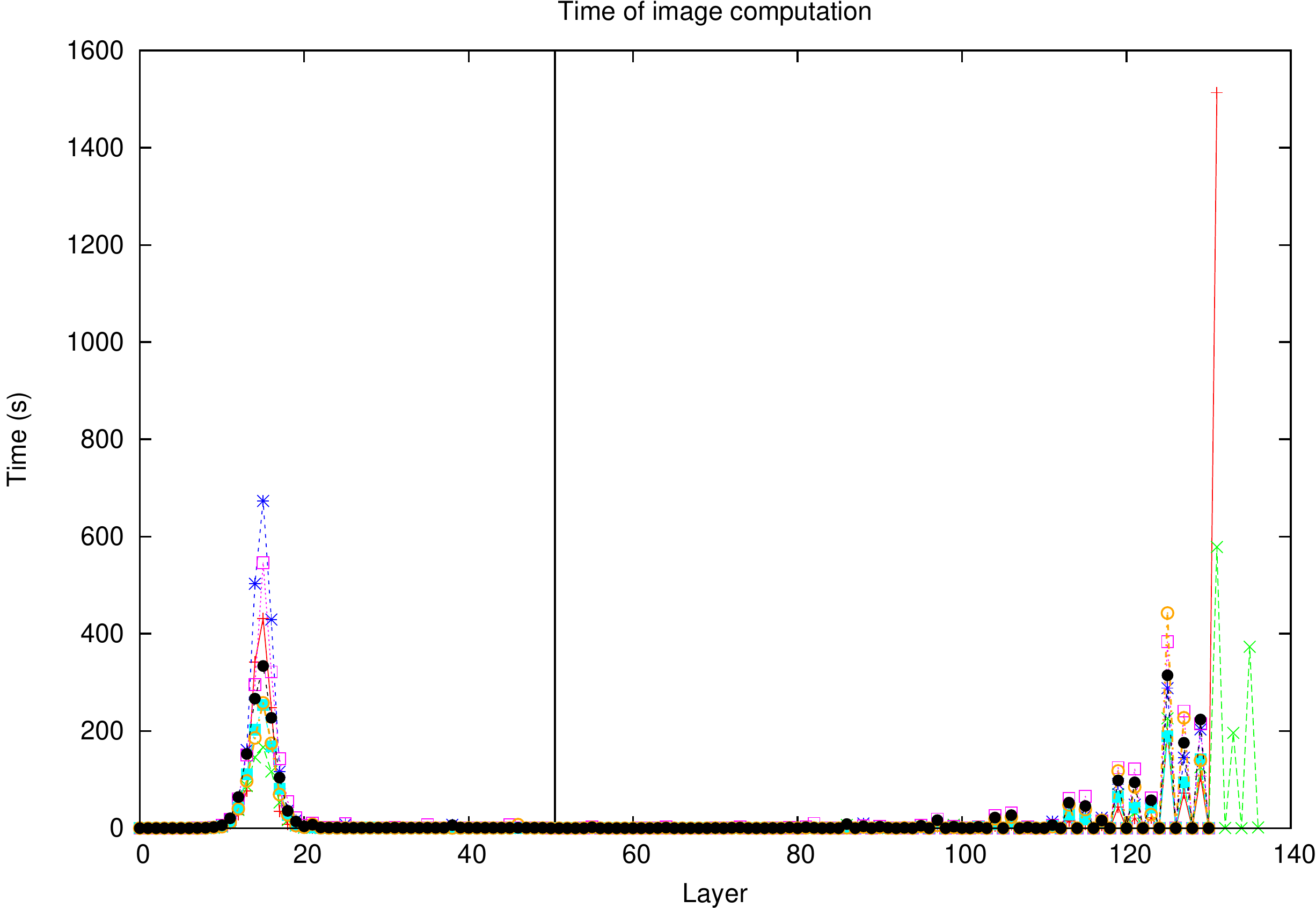}}

\caption[Total time used for each partitioning algorithm in each layer]
{Total time used for each partitioning algorithm in each layer.

\begin{tabular}{rlcrl}
\\
\legnone      & None                  && \legstateslex  & Fold States Lex 8 \\
\legnodescomp & Fold Nodes Compress 8 && \legnodesspath & Fold Nodes Shortest Path 8 \\
\legdisjvar   & Disjunctive Variable  && \legdisjite    & Disjunctive Iterative  \\
\legdisjgen   & Disjunctive Generation &&                & \\
\end{tabular}
}
\label{fig:time}
\end{figure}

\begin{figure}[hbtp]
\centering
\subfloat[8 Puzzle]{\includegraphics[width=7.6cm]{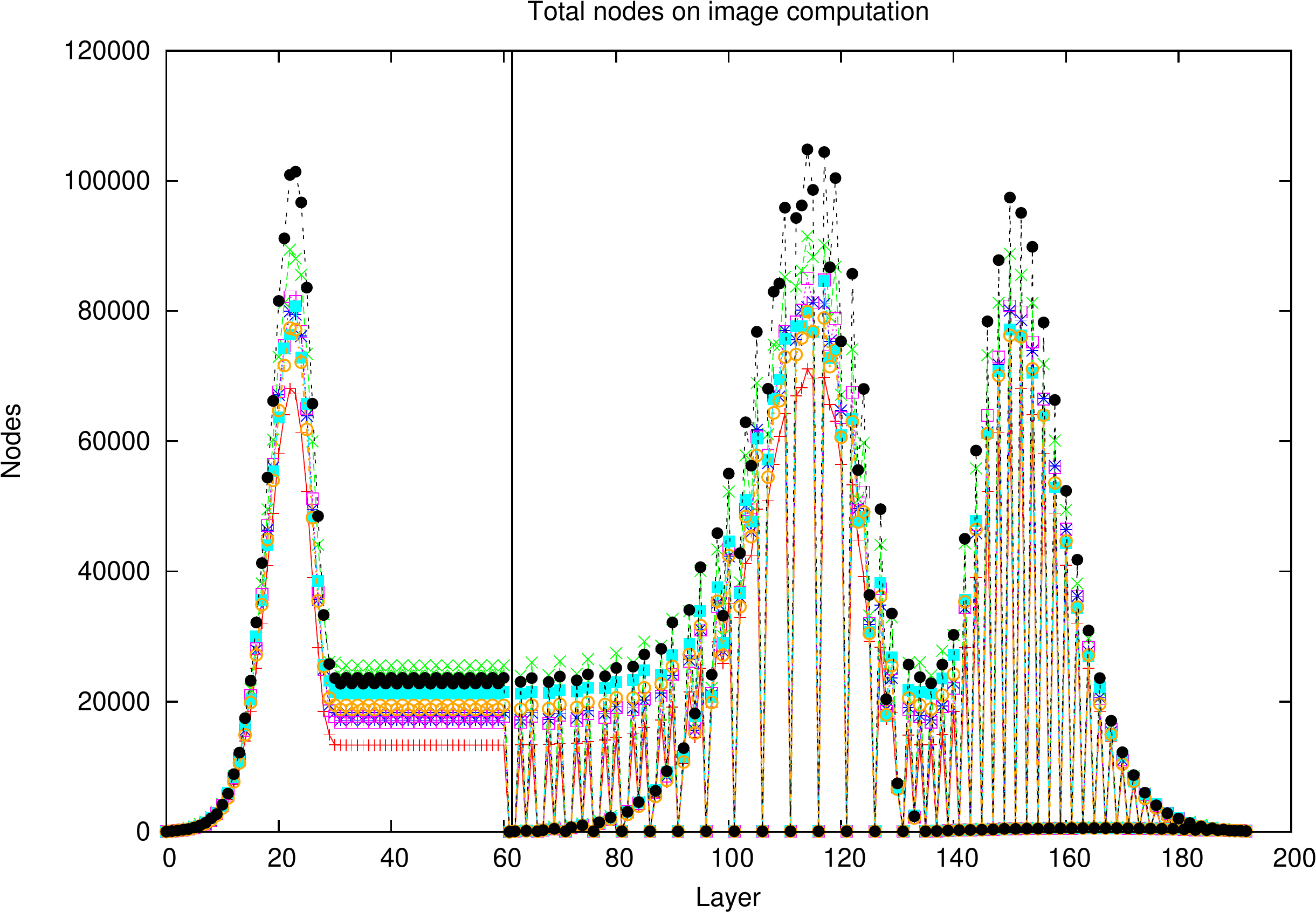}}
\qquad\subfloat[Connect Four $5\times 5$]{\includegraphics[width=7.6cm]{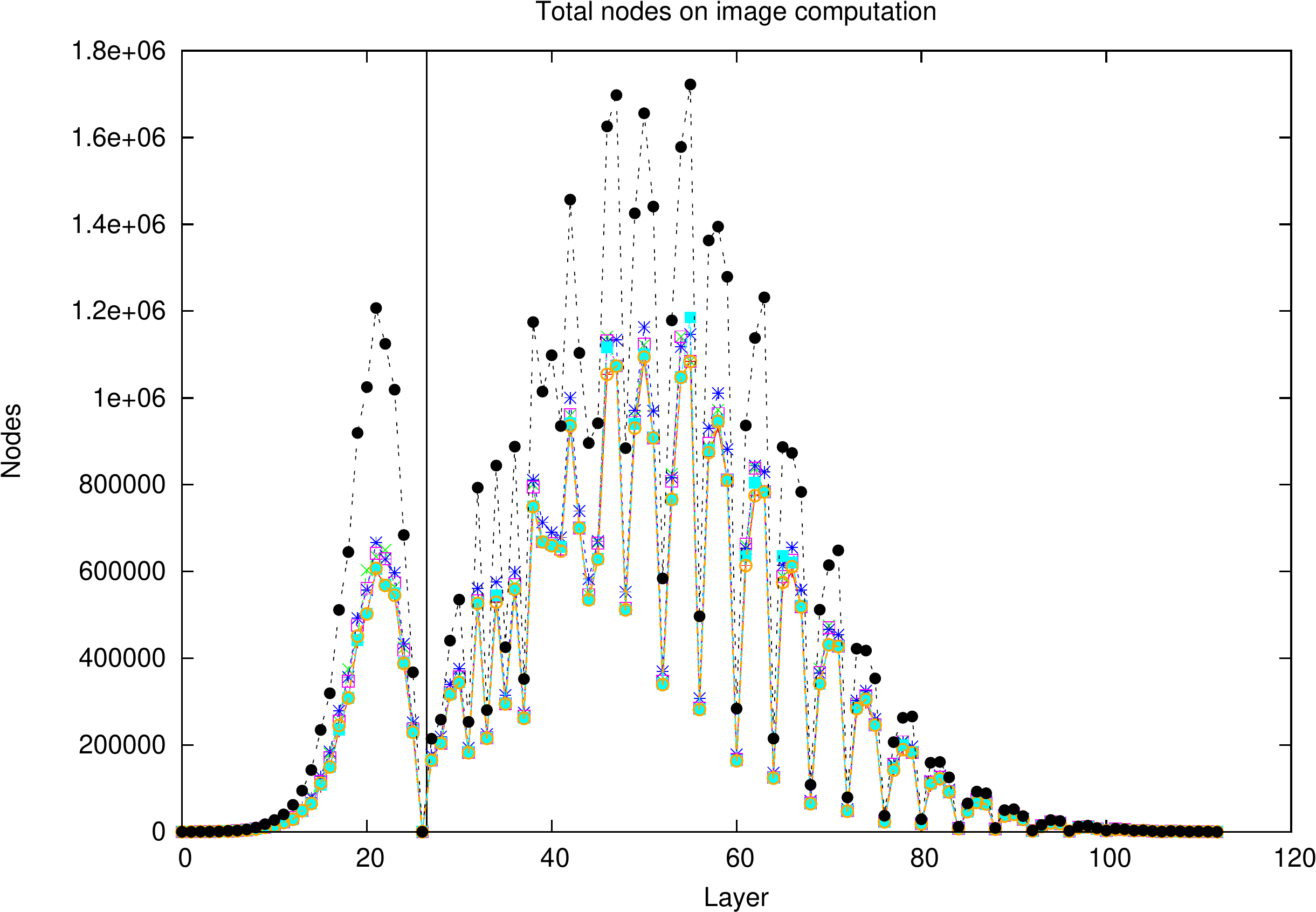}}
\\
\subfloat[Knights Tour]{\includegraphics[width=7.6cm]{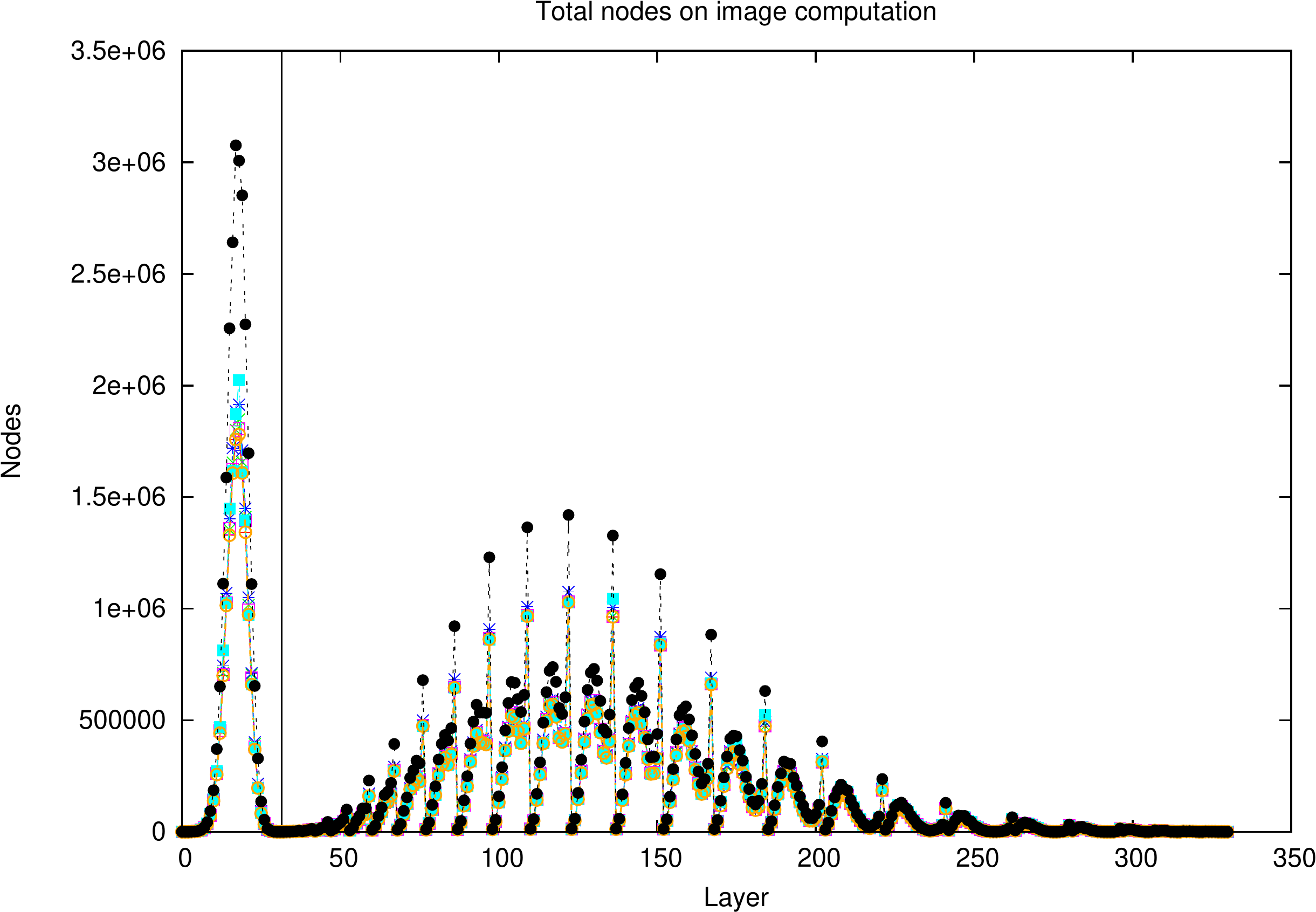}}
\qquad
\subfloat[Lightsout]{\includegraphics[width=7.6cm]{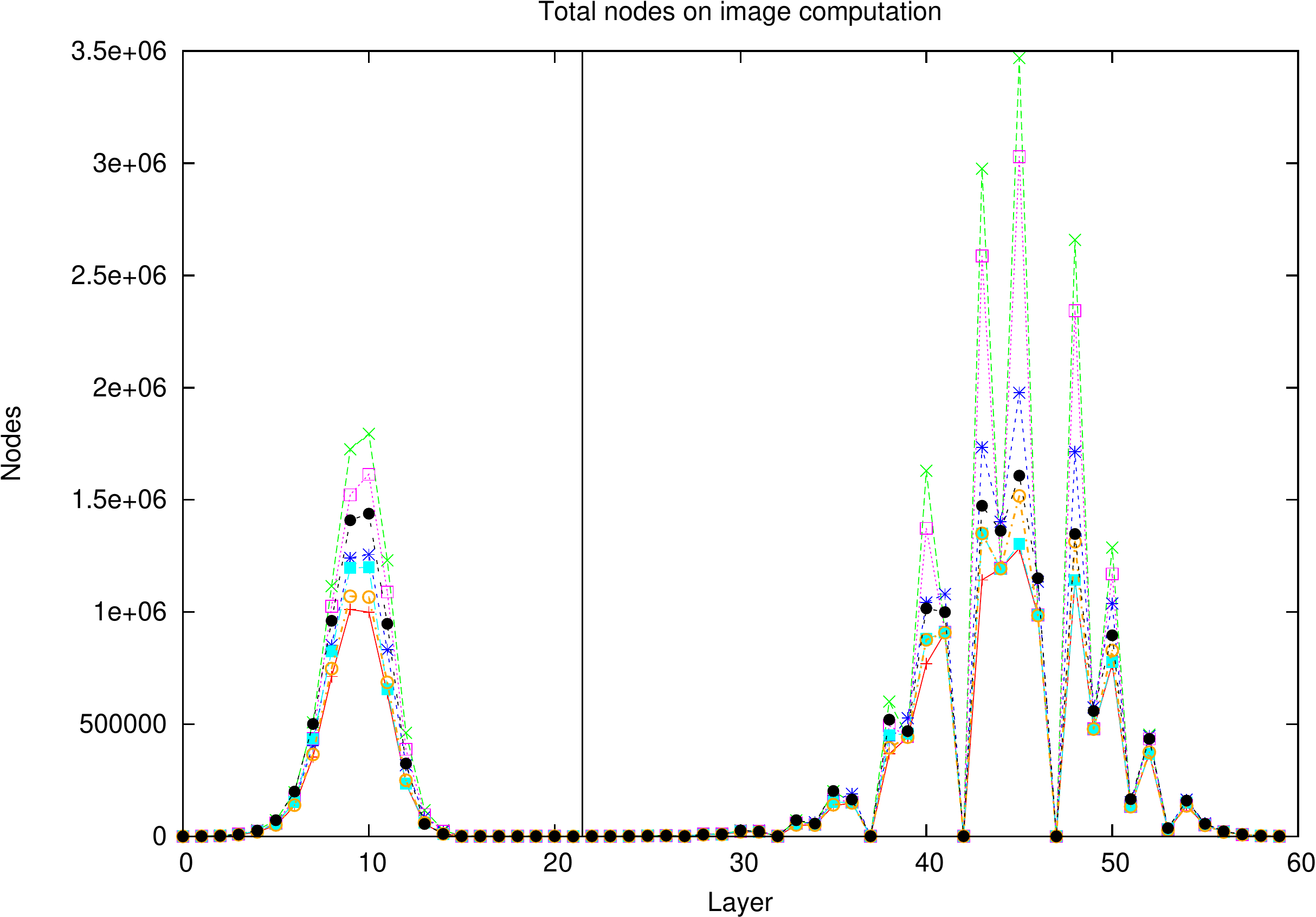}}
\\
\subfloat[TPeg]{\includegraphics[width=7.6cm]{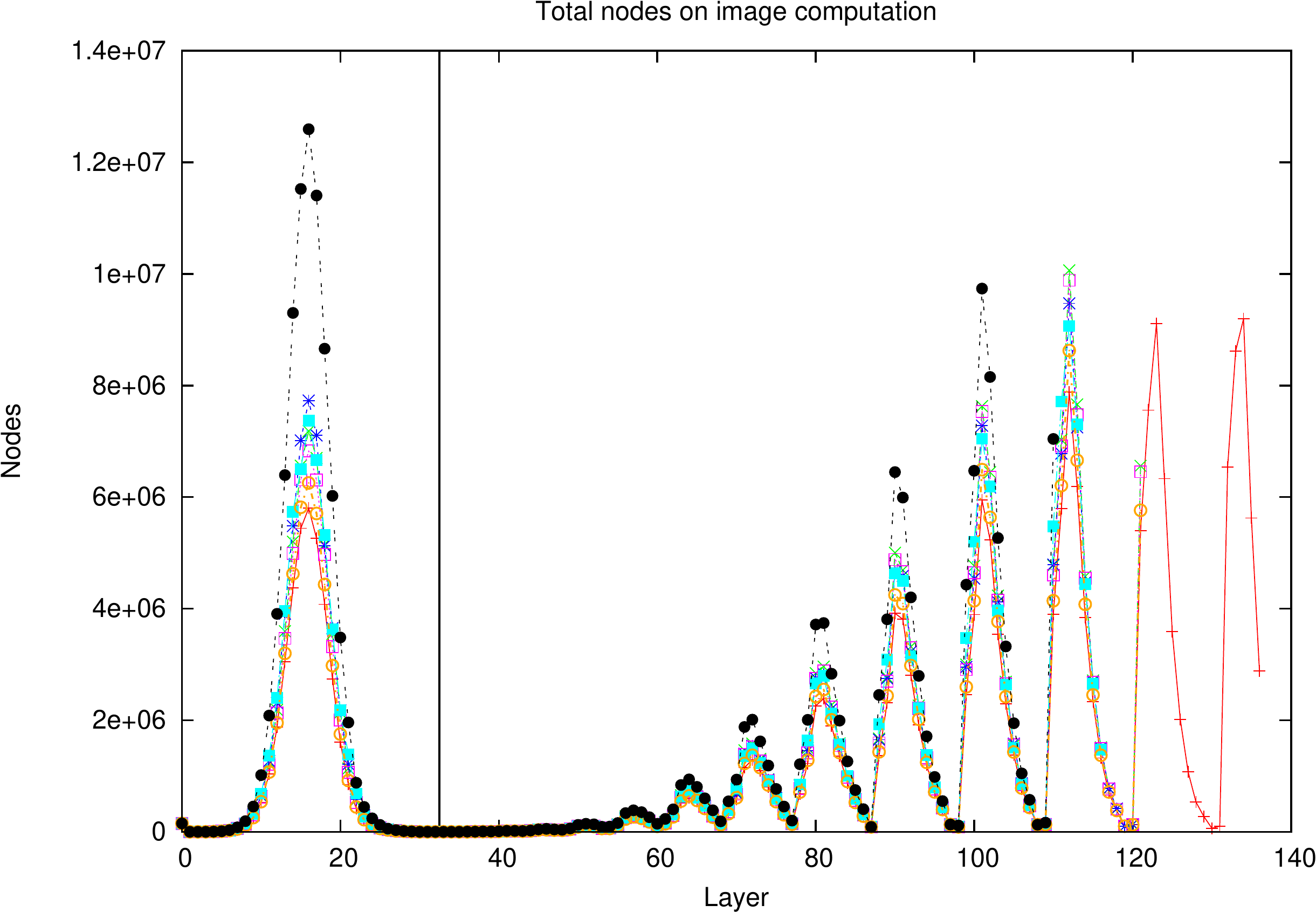}}
\qquad
\subfloat[Asteroids Parallel]{\includegraphics[width=7.6cm]{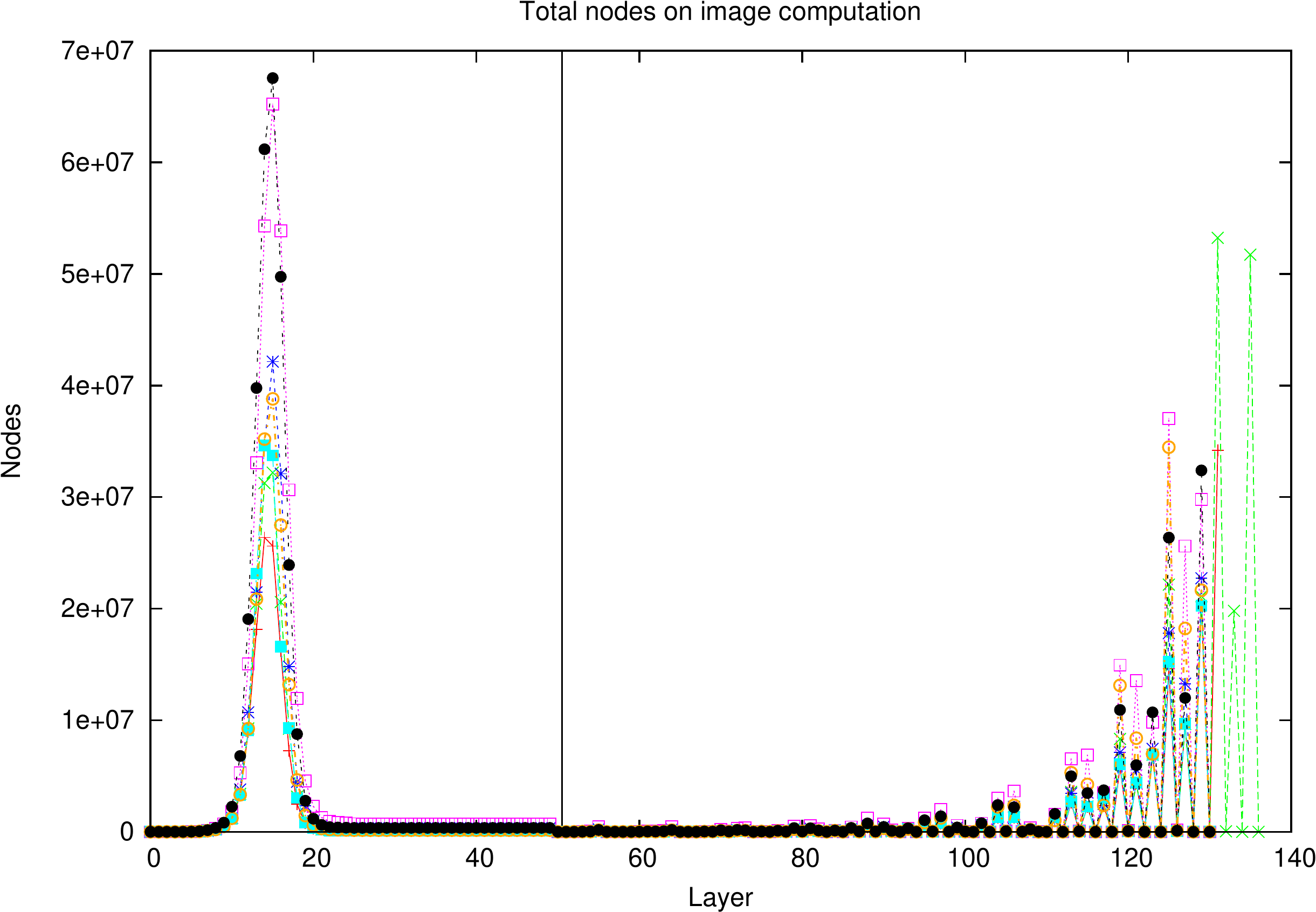}}

\caption[Total number of nodes used for each partitioning algorithm in each layer]
{Total number of nodes used for each partitioning algorithm in each layer.

\begin{tabular}{rlcrl}
\\
\legnone      & None                  && \legstateslex  & Fold States Lex 8 \\
\legnodescomp & Fold Nodes Compress 8 && \legnodesspath & Fold Nodes Shortest Path 8 \\
\legdisjvar   & Disjunctive Variable  && \legdisjite    & Disjunctive Iterative  \\
\legdisjgen   & Disjunctive Generation &&                & \\
\end{tabular}
}
\label{fig:nodes}
\end{figure}

\begin{figure}[hbtp]
\centering
\subfloat[8 Puzzle]{\includegraphics[width=7.6cm]{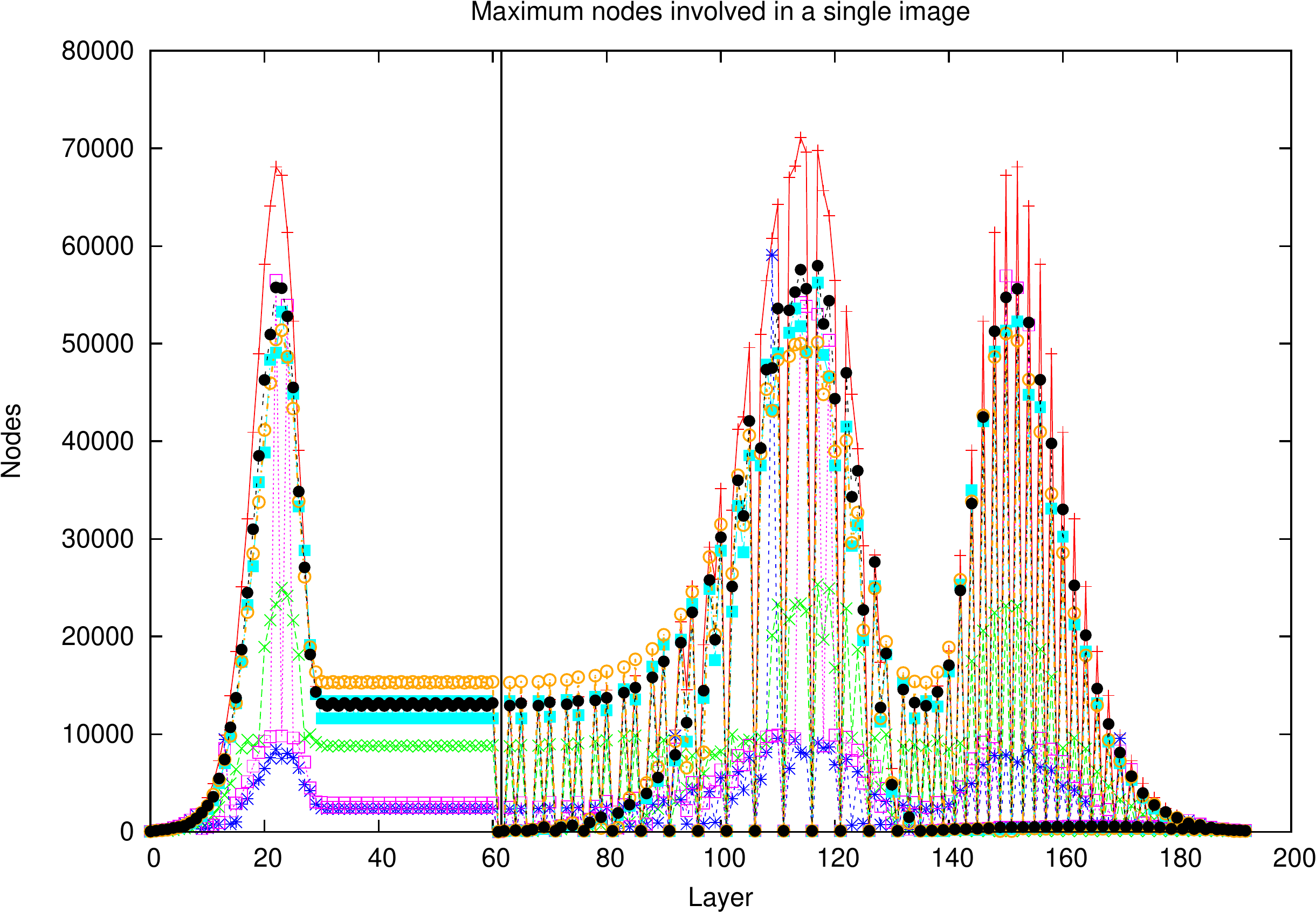}}
\qquad\subfloat[Connect Four $5\times 5$]{\includegraphics[width=7.6cm]{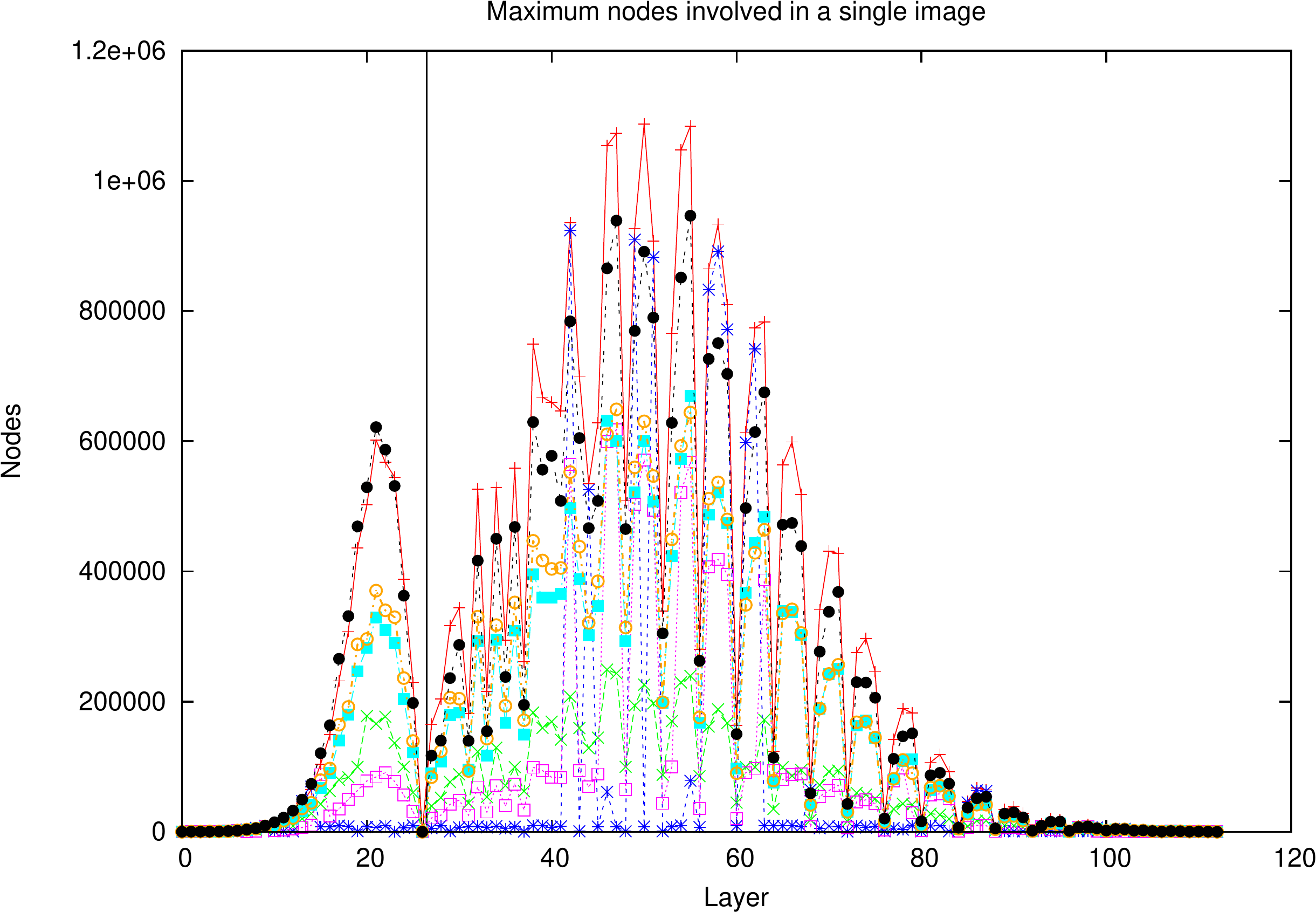}}
\\
\subfloat[Knights Tour]{\includegraphics[width=7.6cm]{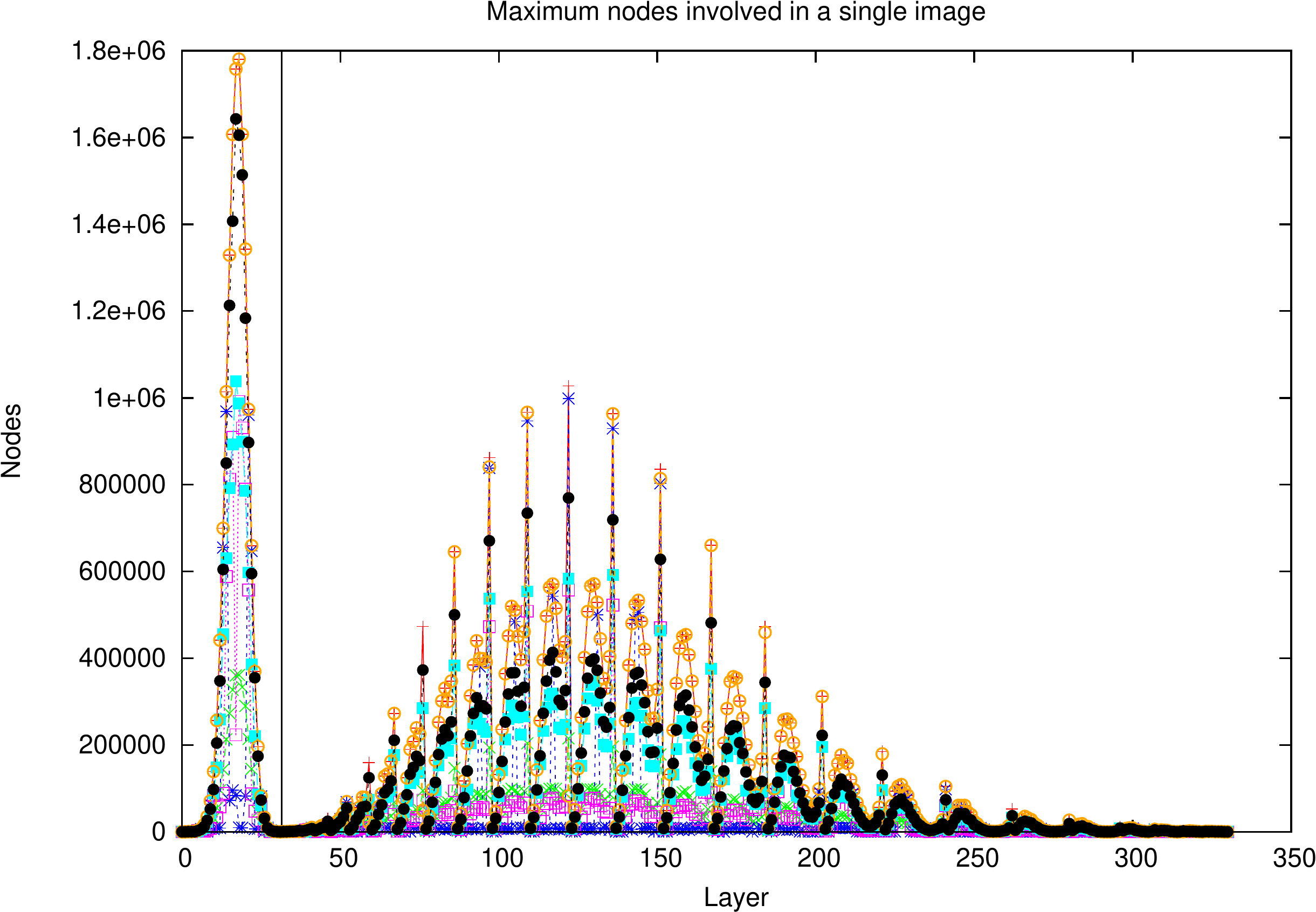}}
\qquad
\subfloat[Lightsout]{\includegraphics[width=7.6cm]{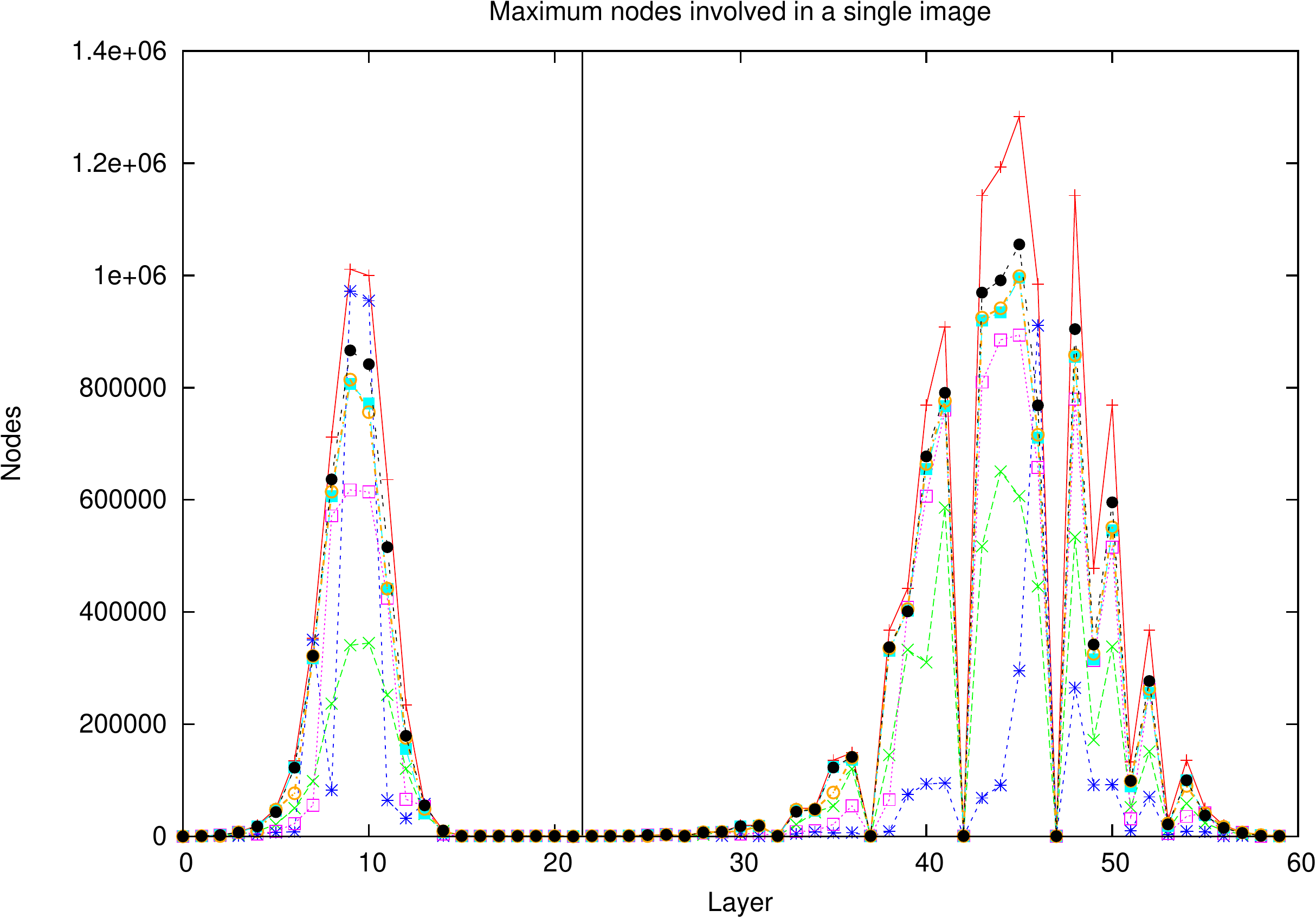}}
\\
\subfloat[TPeg]{\includegraphics[width=7.6cm]{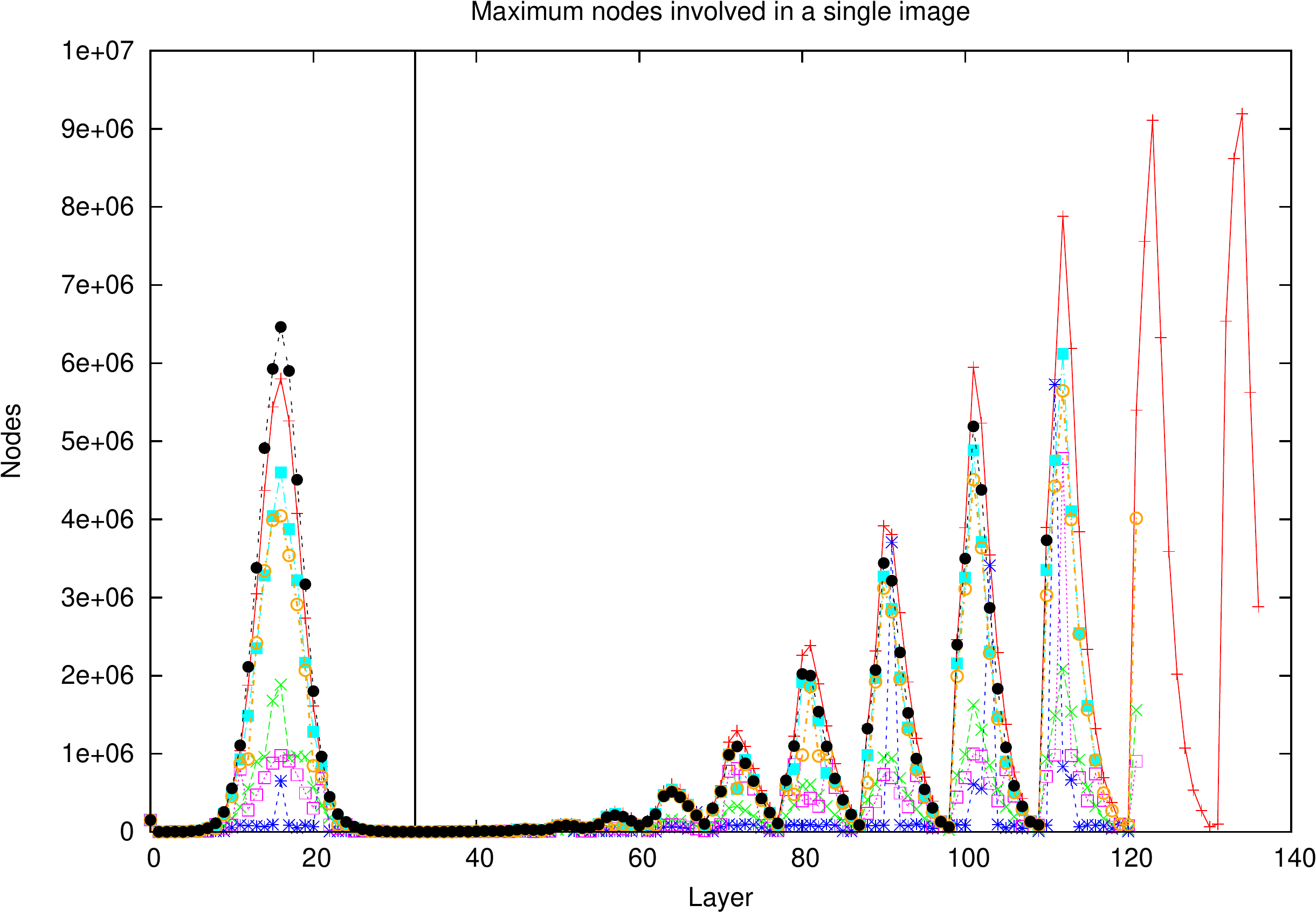}}
\qquad
\subfloat[Asteroids Parallel]{\includegraphics[width=7.6cm]{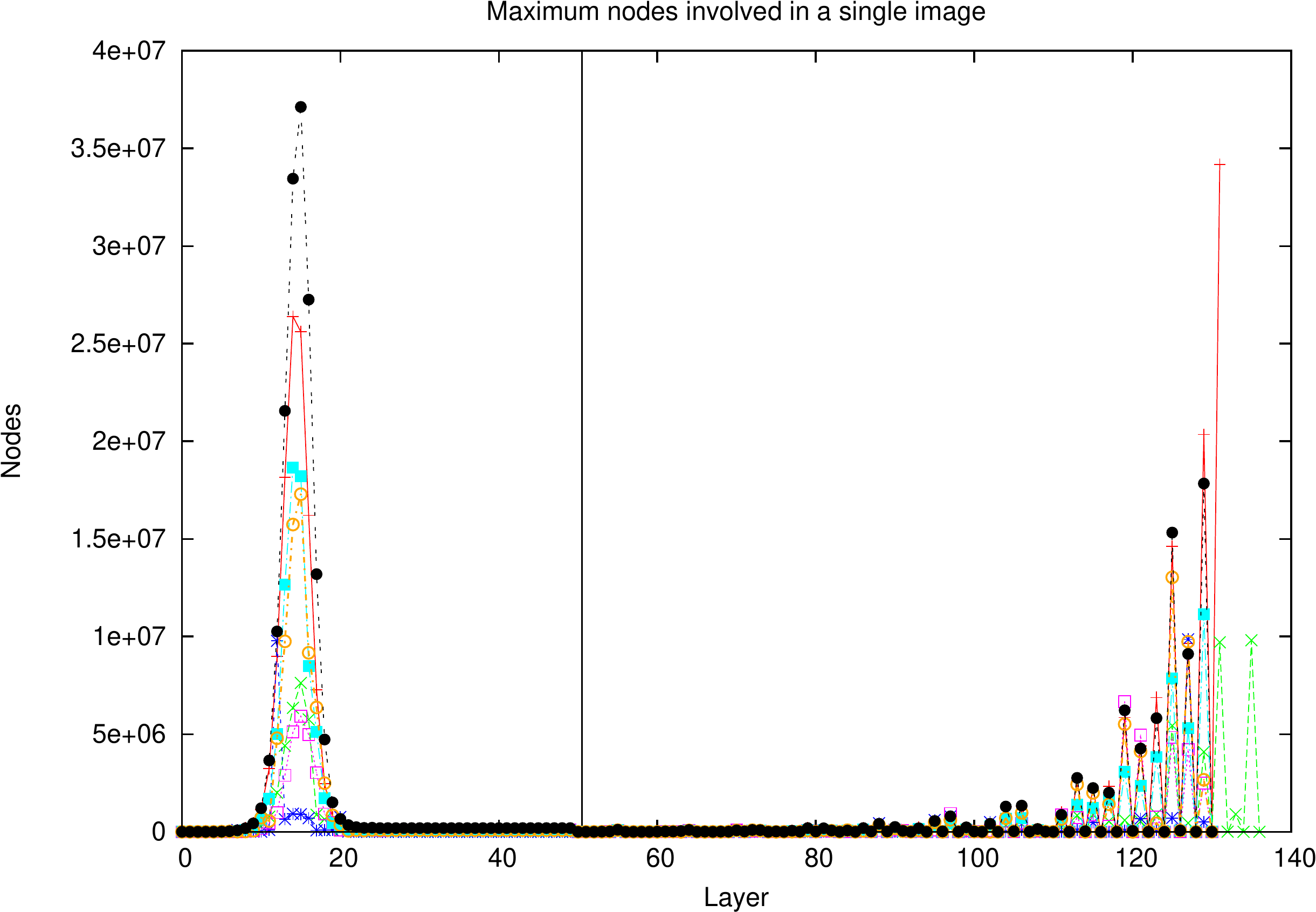}}

\caption[Maximum image size used for each partitioning algorithm in each layer.]
{Maximum image size used for each partitioning algorithm in each layer.

\begin{tabular}{rlcrl}
\\
\legnone      & None                  && \legstateslex  & Fold States Lex 8 \\
\legnodescomp & Fold Nodes Compress 8 && \legnodesspath & Fold Nodes Shortest Path 8 \\
\legdisjvar   & Disjunctive Variable  && \legdisjite    & Disjunctive Iterative  \\
\legdisjgen   & Disjunctive Generation &&                & \\
\end{tabular}
}
\label{fig:image}
\end{figure}

In the plots (Figures~\ref{fig:time}--\ref{fig:image}) we provide
information on the images computed during the exploration in 6 different games of varying difficulty comparing different partitioning options. We omit the methods with less coverage (\emph{States Lex 100000}, \emph{Nodes Shortest Path 10000} and \emph{Nodes Compress 10000}) to visualize well the differences between the other versions.
Both search directions are provided in one plot, separated by a vertical line.

All the selected partitioning options fail to finish the exploration on TPeg and Asteroids Parallel, while they complete the search in all the other domains. In TPeg, the non-partitioning version is the one exploring more layers. On the other hand, in Asteroids Parallel, the lexicographic partitioning achieves the best results.

Figure~\ref{fig:time} shows the time spent in computing each layer by the different partitioning methods. In general, applying partitioning over the BDDs slows the image computation, except in Asteroids Parallel. However, the overhead does not appear prohibitive for the application of partitioning in order to reduce the memory requirements.
The efficiency of the partitioning methods is consistent within each domain, across all the layers. Among the different partitioning options, the \emph{States Lex} version is the closest to the none partitioning across the different domains, followed by \emph {Disjunctive Iterative} and \emph{FoldNodes SPath 8}.

Figure~\ref{fig:nodes} shows the total number of nodes involved in computing each layer, when applying each different method. For all partitioning methods, the sum of the nodes of each part is larger than the original, represented by the none partitioning version.
On the other hand, the total number of nodes of \emph {States Lex} partitioning is not lower than that of the other partitioning methods, suggesting that the improvements in runtime with respect other partitioning methods is thanks to the fast computation of the lexicographic splitting procedure.

Figure~\ref{fig:image} represents the maximum size of the BDDs involved in a single image, determining the minimum memory requirements for performing the task. In this case, applying partitioning seems to help in most cases. Partitionings balancing the number of nodes, \emph{Fold Nodes Compress 8} and \emph{Fold Nodes Shortest Path 8} achieve the best performance in this metric. The lexicographic partitioning is also able to obtain significant reductions in the amount of required memory.

\section{Conclusions and Future Work}
\label{sec:conclusion}

In this paper we have shown an insightful approach for a balanced
disjunctive image computation by splitting the BDD representing the
state set into equally sized subsets to strongly solve games. 
A new BDD-splitting method based on a lexicographic criteria have been presented
and compared to other partitionings in game solving tasks, showing that it is an effective method to reduce the memory requirements. 

Even though we concentrated on game playing in AI the methods we
proposed are general and likely applicable to other areas, such as the
formal verification of sequential circuits. In the model checking
domain BDD partitioning techniques, e.g., by~\cite{Stangier, Emerson},
often show significant advances for executing the image operation, but
usually provide no control on the size of the BDD, nor on the number
of states represented. Moreover, the actual work executed for
computing the image appears to correlate not only with the BDDs in the
input but also with the number of states
represented~\cite{PerformanceBDD}. The core advantage of partitioned
BDDs is that one can gain depth in some of the partitions and locate
errors fast. A complete search of such prioritization is involved, as
information for backtracking has to be maintained. For this case
lexicographic partitioning can help to gain a better search control.
Some potential may be obtained in using different variable orderings
in different partitionings, but this will make the algorithms more
complicated.

Previous work on improved BDD partitioning and guided exploration in
AI includes work by~\cite{EdelkampR98,Jensen08} that partitions the
state sets along different $g$- and $h$-values and along the
difference in the $h$-value. Using a lex-partitioning in the state
sets prior and posterior to computing the image, the search can be
distributed. Each computing node is responsible for computing images
in the lexicographical window it is assigned to. As the images are
computationally expensive, sending around the BDDs is negligible. By
adapting the window sizes different forms of dynamic load balancing
are immediate.


Explicit search can be more space-efficient if perfect hash functions
are available. With ranking and unranking we can eventually connect a
symbolic state space representation with BDDs and an explicit
bitvector based exploration. The BDDs can serve as a basis for a
linear-time ranking and unranking scheme. As we have control over the
number of states in a BDD, we can switch between symbolic and explicit
state space generation when the available main memory is sufficient to
cover the partitioned state sets. This provides a combination of the
two methods, where the BDDs are used to define hash functions for
addressing states in the bitvector representation of the state space.

In game playing we can think of a layered approach to perform forward
search with a BDD and retrograde analysis that changes from symbolic
to explicit-state representation to strongly solve a game, by means
that the solvability status for all states is computed. In case of
exploring domains with complex cost functions in AI planning, a
breadth-first BDD enumeration might be feasible, while computing the
optimal cost is much harder.
  
As the computation of the reachability set is done in compact form the
problem of invalid (unreachable) states in the backward traversal can
be avoided. Due to the partial description of the goals there are many
planning domains where the set of backward reachable states is much
larger than the one in forward search. Consider the sliding-tiles
puzzle with $n$ tiles and with the blank position not mentioned in the
goal state. The inverse of planning operators that move a tile has the
position of the blank and the tile to be moved in the precondition,
and the exchange of tile and blank in the effects. Since the blank
position is not known to the planner backward exploration will
generate states with tiles on top of each other, so that with $n^n$
the set of backward reachable states is exponentially larger than the
number of $n!/2$ forward reachable states.

\paragraph{Acknowledgments}

Thanks to Deutsche Forschungsgemeinschaft (DFG) for support in the
project ED $74/11$-$1$. Thanks to the Spanish Government for support
in the MICINN projects TIN2011-27652-C03-02, TIN2008-06701-C03-03 and
to the Comunidad de Madrid for support in the project
CCG10-UC3M/TIC-5597.

\bibliography{conn4}
\bibliographystyle{eptcs}
\end{document}